\documentclass{article} 
\usepackage{iclr2027_conference,times}

\usepackage{amsmath,amsfonts,bm}

\def\eqref#1{equation~\ref{#1}}

\def\1{\bm{1}}

\DeclareMathAlphabet{\mathsfit}{\encodingdefault}{\sfdefault}{m}{sl}
\SetMathAlphabet{\mathsfit}{bold}{\encodingdefault}{\sfdefault}{bx}{n}

\usepackage[T1]{fontenc}
\usepackage{hyperref}
\usepackage{url}
\usepackage{amsmath}
\usepackage{amssymb}
\usepackage{bbm}
\usepackage{graphicx}
\usepackage{booktabs}
\usepackage{multirow}
\usepackage{wrapfig}
\usepackage{subcaption}
\usepackage{tcolorbox}

\title{Capacity, Responsiveness and Alignment: \\ What Makes a Latent Structure Actionable}

\author{
Or Shafran \qquad Mor Geva \\
Blavatnik School of Computer Science and AI, Tel Aviv University, Israel
}

\author{Or Shafran \qquad \textbf{Mor Geva} \vspace{3pt}\\
Blavatnik School of Computer Science and AI, Tel Aviv University \vspace{3pt}\\
\texttt{ordavids1@mail.tau.ac.il}
}
\iclrfinalcopy
\usepackage{bm}

\begin{document}

\maketitle
\lhead{Preprint}

\begin{abstract}

Localizing latent structures in the activation space of language models (LMs) is central to understanding and controlling their behavior. Yet, localized structures can differ substantially in their causal influence, raising the question of what makes a structure actionable. We tackle this question by casting causal influence as a product of three factors and showing empirically that they act as interpretable, distinct constraints: \textit{capacity}, measuring the sensitivity of the model's output to movement along the structure, \textit{responsiveness}, capturing how promotable the concept is given the current context, and \textit{alignment}, reflecting how well the structure aligns with the context-specific representation of the concept. Across 4 LM families and 50 concepts, we observe that causal effectiveness requires all factors to be high; low capacity and responsiveness reduce it by 84\% and 95\%, respectively, while low alignment can reverse it,  suppressing concept expression. Moreover, we find that causality is context-dependent rather than an intrinsic property of the structure, with causally effective directions forming a low-dimensional subspace that varies across contexts. By restricting the training of linear probes to this subspace, we introduce \textit{causal probes} that achieve 17\%--118\% improvement in steering across models, with only 3\% reduction in concept detection.
\end{abstract}

\section{Introduction}

A standard interpretability approach for controlling language models (LMs) is to localize geometric structures in activation space, such as directions, subspaces, or manifolds, that are associated with specific concepts or behaviors \citep[][\textit{inter alia}]{belinkov-etal-2017-neural, conneau-etal-2018-cram, hewitt-manning-2019-structural, turner2024steeringlanguagemodelsactivation, huben2024sparse, park2024linear, sun2025hyperdas, huang2026decomposing, tiblias2026hypothesis, wurgaft2026manifold} and then intervene along these structures. This approach rests on the assumption that if a structure can be \textit{localized} in hidden activations then intervening on it should reliably induce the intended change in the model's output distribution.

Despite its wide adoption, this assumption often breaks in practice. A localized structure may only reflect correlations in the data \citep{ravfogel2020null, elazar2021amnesic, belinkov2022probing}, and even when it is causally relevant, interventions along such structure can produce weak, inconsistent, or unintended effects \citep{arad2025saes, wu2025axbench}. Moreover, recent work shows that directions optimized for detection often differ from those that causally influence model outputs \citep{galeone2026perfectdetectionfailedcontrol}. Together, these findings establish that localization alone is insufficient for causal control, raising the question of what makes a latent structure actionable.

In this work, we tackle this question by decomposing the causal influence of a latent structure at individual contexts. Given a structure $\mathcal{S}$ and activation $\mathbf{h}$, we ask how much an intervention along $\mathcal{S}$ at $\mathbf{h}$ can increase the expression of a target concept $c$.
Here, we define concept expression as the probability mass assigned to tokens associated with the concept, selected from the vocabulary $\mathcal{V}$ via an indicator vector $\mathbf{r}_c\in\{0,1\}^{|\mathcal{V}|}$. For an output distribution $\mathbf{p}$, the expression of $c$ is therefore $\mathbf{r}_c^\top\mathbf{p}$.
We then analyze how much an intervention along $\mathcal{S}$ at $\mathbf{h}$ can change this quantity. Let $\mathbf{v}$ denote an intervention direction and $J_{\mathbf{h}}$ the Jacobian mapping changes in $\mathbf{h}$ to changes in the output logits. The resulting first-order change in concept expression can be cast as a product of three factors:
\begin{equation*}
\underbrace{\sigma_{\mathbf{p}}(J_{\mathbf{h}}\mathbf{v})}_{\text{\text{\textcolor{cyan}{I}}}}
\underbrace{\sigma_{\mathbf{p}}(\mathbf{r}_c)}_{\text{\textcolor{cyan}{II}}}
\underbrace{\operatorname{Corr}_{\mathbf{p}}(\mathbf{r}_c,J_{\mathbf{h}}\mathbf{v})}_{\text{\text{\textcolor{cyan}{III}}}}
\end{equation*}
where 
$\sigma_{\mathbf{p}}(\cdot)$ and $\operatorname{Corr}_{\mathbf{p}}(\cdot)$ are the standard deviation and correlation under $\mathbf{p}$, respectively.

Next, we show empirically that these factors capture distinct and interpretable constraints on causal influence.
Factor \textcolor{cyan}{I} captures \textbf{causal capacity}, signifying whether changes along $\mathcal{S}$ at $\mathbf{h}$ can propagate through the model's downstream computation to generally affect its output. We find that this sensitivity to local changes in activation space is concentrated in a context-dependent, low-rank subspace.
Factor \textcolor{cyan}{II} encodes \textbf{concept responsiveness}, capturing the extent to which the concept is responsive to interventions under the given context. We observe that finite interventions primarily promote concept-relevant tokens that already receive non-negligible probability, with the top $2.5\%$ of tokens under $\mathbf{p}$ receiving $94.5\%$ of the intervention-induced probability increase. Last, factor \textcolor{cyan}{III} corresponds to \textbf{concept alignment}, capturing whether intervening along $\mathcal{S}$ at $\mathbf{h}$ is correctly oriented to promote the concept. We find that the directions promoting the concept vary substantially across contexts, 
where the same intervention direction can be well aligned in one context and poorly aligned in another.
Importantly, reducing any of the factors bottlenecks the causal effect of $\mathcal{S}$, reducing it by 84--95\% when one of the factors is low and driving it to zero when all are low. Low alignment can even cause the same latent structure to suppress its associated concept.

We then demonstrate the practical utility of our findings for steering. We show that linear probes often capture directions with little causal effect. By restricting the learned directions to high capacity subspaces, we introduce \textit{causal probes} which improve the median steering performance by \(17\%\)--\(118\%\) across models, while reducing concept-detection accuracy by only \(3\%\).

Overall, our results show that the causal influence of a latent structure is not an intrinsic property of the structure itself, but rather arises from the relationship between the structure, context, and remaining computation. Capacity, responsiveness, and alignment are distinct factors describing this relationship, together determining to what extent a concept's expression can be promoted from the current state through the latent structure. We release our code at \url{https://github.com/ordavid-s/what-makes-a-latent-causal}.

\section{Disentangling Causal Influence}
\label{method}

We study what determines the ability of a latent structure to causally induce the expression of a concept. To this end, we quantify concept expression at the model's output and pull it back to intermediate activations. Then, we analyze the causal effect of movement along the structure at a given activation, decomposing it into three constraints.

\paragraph{Pulling back causal influence}

Assume an auto-regressive LM with vocabulary $\mathcal{V}$. For an activation $\mathbf{h}\in\mathbb{R}^d$ at an intermediate layer, we denote by $\mathbf{z}$ the output logits produced by the remaining layers, and by $\mathbf{p}=\operatorname{softmax}(\mathbf{z})$ the corresponding probability vector over $\mathcal{V}$. Since the causal effect of a latent structure ultimately manifests as a change in the model's output distribution (at the current or future positions), we measure a concept's expression by the probability mass assigned to tokens that verbalize it.
For example, the concept of refusal should promote tokens such as \texttt{cannot} and \texttt{won't}. Formally, we encode a set of tokens $\mathcal V_c\subseteq\mathcal V$ that verbalize a concept $c$ as a vector:\footnote{We describe how we systematically construct such concept token sets in \S\ref{experimental_setting}.}
\begin{equation}
(\mathbf{r}_c)_i =
\begin{cases}
1, & i\in\mathcal V_c,\\
0, & \text{otherwise}.
\end{cases}
\label{eq:indicator}
\end{equation}
For a change $\Delta\mathbf{p}$ in the output distribution, the measured effect on concept expression is simply $\mathbf{r}_c^\top\Delta\mathbf{p}$. Thus, $\mathbf{r}_c^\top$ acts as an output-space covector\footnote{A covector is a linear map from vectors to scalars.} that measures whether a local change in the distribution promotes or suppresses the concept.

To quantify the causal effect of a latent structure $\mathcal{S}$ on $c$, we consider how intervening along $\mathcal{S}$ changes concept expression. For an intervention vector $\mathbf{v}$ along $\mathcal{S}$ at $\mathbf{h}$, we obtain a distribution change $\Delta \mathbf{p} = f(\mathbf{h} + \mathbf{v}) - f(\mathbf{h})$, where $f(\cdot)$ denotes the remaining computation from $\mathbf h$ to the output.
The first-order change in the output logits is
\begin{equation}
\Delta\mathbf{z} \approx J_{\mathbf{h}}\mathbf{v}, \qquad J_{\mathbf{h}}=\partial\mathbf{z}/\partial\mathbf{h},
\end{equation}
and passing this change through the softmax gives the first-order change in the output distribution,
\begin{equation}
 \Delta\mathbf{ p} \approx F_{\mathbf{p}} J_{\mathbf{h}} \mathbf{v},
\qquad
F_\mathbf{p}=\operatorname{diag}(\mathbf{p})-\mathbf{p}\mathbf{p}^\top,
\label{eq:fisher_info_matrix}
\end{equation}
where $F_{\mathbf{p}}$ is the Jacobian of softmax. Since concept expression is measured by $\mathbf r_c^\top \Delta \mathbf p$, the resulting first-order change is
\begin{equation}
\mathbf{r}_c^\top \Delta \mathbf{p}
\approx
\mathbf{r}_c^\top F_{\mathbf{p}} J_{\mathbf{h}} \mathbf{v}.
\label{eq:causal-effect}
\end{equation}
Thus, $\mathbf r_c^\top F_{\mathbf p}J_{\mathbf h}$ pulls the output-space concept covector $\mathbf r_c^\top$ back to the intermediate activation space, characterizing which local changes to $\mathbf h$ affect concept expression. We evaluate the quality of this first-order approximation in \S\ref{sec:capacity_interp}.

\paragraph{Three factors construct causal influence}
Having quantified the causal effect of movement along $S$ on the output, we now ask what determines this effect. Using the form of the softmax Jacobian in Eq.~\ref{eq:fisher_info_matrix}, the first-order change in concept expression can equivalently be written as
\begin{equation}
\mathbf{r}_c^\top \Delta \mathbf{p}
\approx
\operatorname{Cov}_{\mathbf{p}}
\left(
\mathbf{r}_c,
J\mathbf{v}
\right),
\end{equation}
where $\operatorname{Cov}_{\mathbf{p}}$ denotes the covariance over vocabulary entries weighted by $\mathbf{p}$. Using the covariance–correlation identity,

we cast the first-order effect as three factors:
\begin{equation}
\operatorname{Cov}_{\mathbf{p}}
\left(
\mathbf{r}_c,
J\mathbf{v}
\right)
=
\underbrace{\sigma_{\mathbf{p}}(J_{\mathbf{h}}\mathbf{v})}_{\text{\text{\textcolor{cyan}{I}}}}
\underbrace{\sigma_{\mathbf{p}}(\mathbf{r}_c)}_{\text{\textcolor{cyan}{II}}}
\underbrace{\operatorname{Corr}_{\mathbf{p}}(\mathbf{r}_c,J_{\mathbf{h}}\mathbf{v})}_{\text{\text{\textcolor{cyan}{III}}}}
\label{eq:causal-effect-decomposition}
\end{equation}
Together, these terms form the first-order effect of movement along $\mathcal{S}$. In the next sections, we show that they reveal interpretable (\S\ref{interp_factors}) and distinct (\S\ref{buckets}) constraints on causal influence.

\paragraph{Extending to sequence-level causality} Some latent structures may not directly affect the next-token prediction, but still influence outputs at later positions. To capture this, we extend the first-order effect over a horizon of $T$ positions. Let $J_\mathbf{h}^t=\partial\mathbf{z}_t/\partial\mathbf{h}$ denote the Jacobian from the intermediate activation to the logits at position $t$, and let $\mathbf{p}_t$ denote the corresponding output distribution. The cumulative effect on concept expression is then \begin{equation} \sum_{t=0}^{T}\mathbf{r}_c^\top\Delta\mathbf{p}_t \approx \sum_{t=0}^{T} \mathbf{r}_c^\top F_{\mathbf{p}_t}J_\mathbf{h}^t\mathbf{v}. \end{equation}
Since differentiation distributes over sums, the covector corresponding to the sequence-level effect is simply the sum of the position-wise pullbacks. Each position is therefore characterized through its corresponding $J_\mathbf{h}^t$ and $\mathbf{p}_t$ and
admits the same decomposition as in Eq.~\ref{eq:causal-effect-decomposition}. Thus, we can capture both the intervention's effect at each output position and how these effects accumulate during generation.

\section{Causality is Mediated by Interpretable Factors}
\label{interp_factors}
We showed that the first-order causal effect of 
a latent structure can be cast into three factors (Eq.~\ref{eq:causal-effect-decomposition}). In this section, we show these factors capture distinct, interpretable constraints on causal influence, corresponding to causal \textit{capacity} (\textcolor{cyan}{I}), concept \textit{responsiveness} (\textcolor{cyan}{II}), and concept \textit{alignment} (\textcolor{cyan}{III}).

\subsection{Experimental Setup}
\label{experimental_setting}

\paragraph{Data}
We collect a set of 50 diverse concepts, such as expressions of trust and geographic locations, sampled from a large list of concepts previously identified in LM representations \citep{lieberum2024gemmascopeopensparse}. Then, using a set of 304 open-ended prefixes,
we prompt gpt-4.1-mini \citep{openai2024gpt4ocard} to construct two datasets per concept: \textit{positive} with generations that exemplify the concept, and \textit{neutral} with neutral completions that do not. 
Prompts, example prefixes and more details are in \S\ref{app:data}.
We use these datasets to find latent structures and form the output covectors.

\paragraph{Constructing concept covectors and latent structures}
To compute the three factors, we construct for each concept an output-space covector $\mathbf{r}_c^T$ that measures concept expression and a latent structure $\mathcal{S}$ representing the concept. We construct $\mathbf{r}_c^T$ as detailed in Eq.~\ref{eq:indicator}, by assigning a value of $1$ to tokens that occur \textit{only} in the positive dataset and $0$ otherwise (see \S\ref{app:covectors} for details). For latent structures, we use Difference-in-Means (DiffMeans) \citep{li2023inference, marks2024the} and linear probes \citet{alain2017understanding, hewitt2019designing}, following the formulation of \citet{wu2025axbench}. Both methods are trained using the concept's positive and neutral datasets. We construct these structures independently at each evaluated model layer and intervene along them via the commonly used additive intervention \citep{subramani2022extracting, li2023inference, turner2024steeringlanguagemodelsactivation}, $
\mathbf{h}' = \mathbf{h} + \alpha \mathbf{v}
$, where $\mathbf{v}$ is a direction along $\mathcal{S}$ and $\alpha \in \mathbb{R}$ controls the intervention strength.

\paragraph{Models}
We conduct our experiments across four LMs: Gemma-2-9B, Llama-3.1-8B, Qwen-2.5-7B, and Mistral-7B \citep{gemmateam2024gemma2improvingopen, grattafiori2024llama3herdmodels, jiang2023mistral7b, qwen2025qwen25technicalreport}. To capture how causal influence changes over the course of computation, we evaluate activations at layers corresponding to approximately $20\%$, $40\%$, $60\%$, $80\%$, and $95\%$ of model depth.

\begin{figure}[t]
\setlength{\belowcaptionskip}{-10pt}
    \centering
    \includegraphics[width=0.85\linewidth]{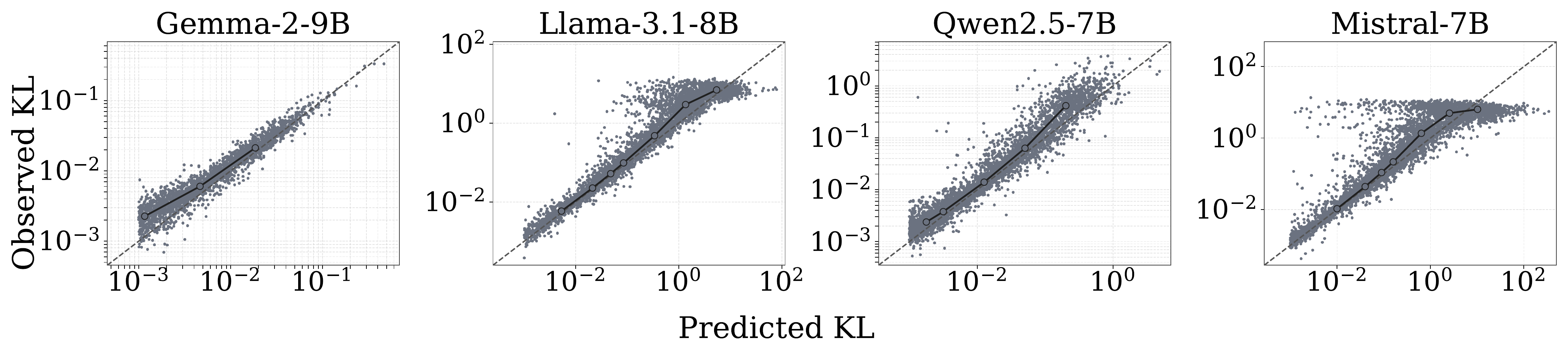}
    \caption{\textbf{Capacity predicts intervention effects.}
    Predicted and observed intervention-induced KL closely agree across intervention strengths.}
    \label{fig:predicted_vs_actual_kl}
    
\end{figure}

\subsection{Capacity: Downstream Sensitivity is Concentrated in a Context-Dependent Subspace}
\label{sec:capacity_interp}

Downstream layers may be highly sensitive to certain changes in intermediate activations while being invariant to others. Factor $\text{\textcolor{cyan}{I}}$, defined by $\sigma_{\mathbf p}(J_{\mathbf h}\mathbf v)$, captures this sensitivity: the activation subspace of directions to which the downstream computation is most sensitive is
\begin{equation}
G_{\mathbf h}
=
J_{\mathbf h}^{\top}F_{\mathbf p}J_{\mathbf h},
\label{eq:G}
\end{equation}
where sensitivity along $\mathbf v$ is
$\sigma_{\mathbf p}(J_{\mathbf h}\mathbf v)
=
\sqrt{\mathbf v^\top G_{\mathbf h}\mathbf v}$.
The factor is large when movement along $\mathbf v$ propagates to produce variation across the output logits, and small when that movement has little downstream effect. We therefore interpret this factor as the causal \emph{capacity} of movement along the structure at the current activation.

\paragraph{Capacity predicts magnitude of intervention-induced output change} 
We test our interpretation by checking whether capacity predicts the magnitude of intervention-induced output change. For activation $\mathbf h$ and output position $t$ (where $t=0$ denotes the current prediction), an intervention $\alpha\mathbf v$ induces a local KL divergence approximated to second order by
$
\frac{\alpha^2}{2}\mathbf v^\top G_{\mathbf h}^t\mathbf v,
$
providing a direct prediction of the resulting output change (derivation in \S\ref{app:kl_derivation}). We compare this predicted KL with the observed KL for interventions along 64 isotropic unit directions over all neutral contexts, using 14 intervention strengths $\alpha$ ranging from $0.3$ to $128$.
Figure~\ref{fig:predicted_vs_actual_kl} shows the results for the current prediction ($t=0$) and depth 60\%, with additional offsets and layers reported in \S\ref{app:capacity_offsets}. Predicted KL closely tracks observed KL, with directions outside the high-capacity subspace producing little causal effect.

\begin{figure}[t]
\setlength{\belowcaptionskip}{-10pt}
    \centering
    \includegraphics[width=0.45\linewidth]{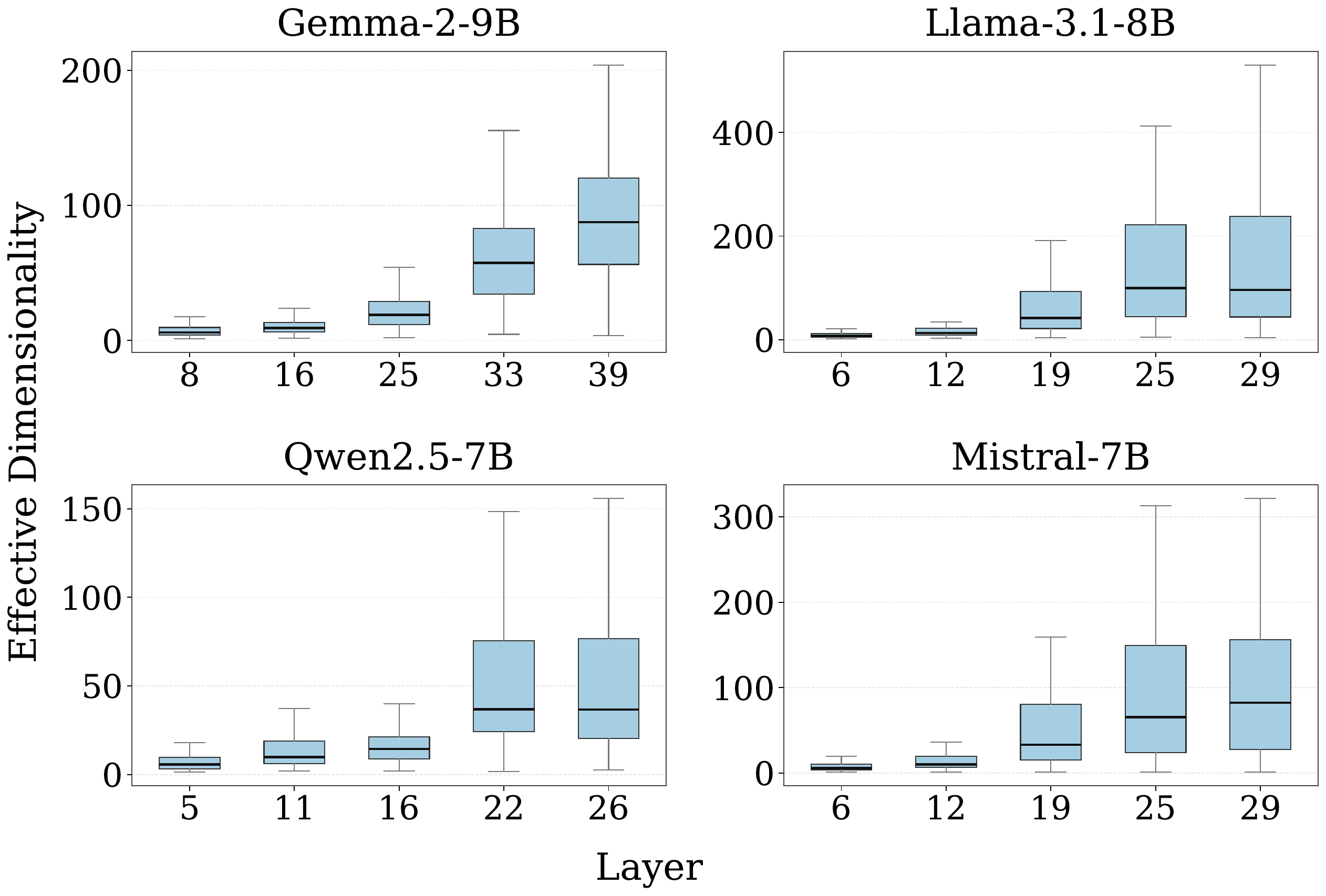}
    \hfill
    \includegraphics[width=0.48\linewidth]{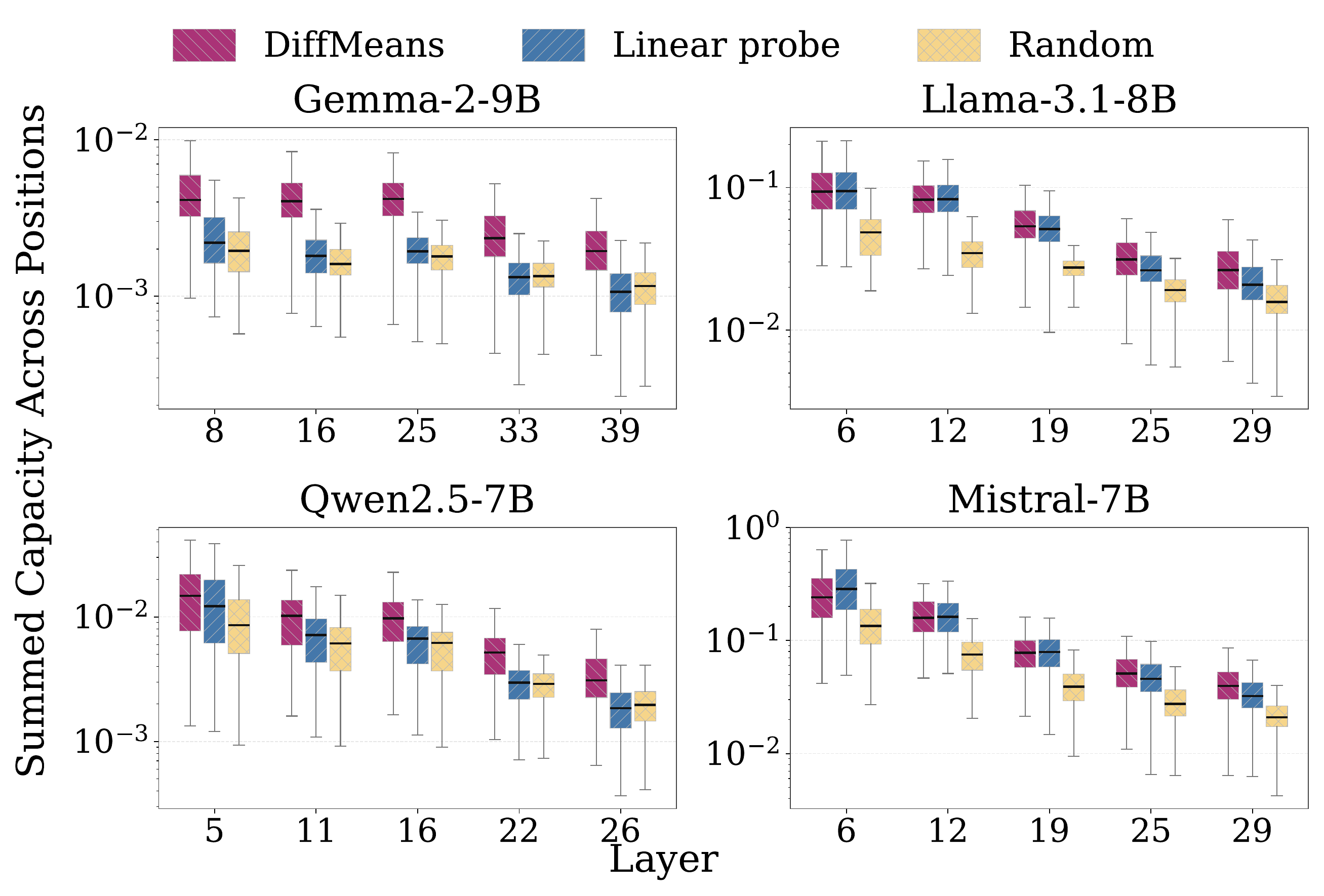}

    \caption{\textbf{Capacity is concentrated in a low-dimensional subspace and varies across latent structures.}
    \textbf{Left:} Effective dimensionality across layers, showing that downstream sensitivity is concentrated in few activation-space directions.
    \textbf{Right:} Capacity of DiffMeans, linear probes and random directions for current and future positions substantially varies between localization methods.}
    \label{fig:capacity_extra}
\end{figure}

\paragraph{Capacity is context dependent and low rank}
Having established that capacity predicts output change, we characterize how this sensitivity is realized in activation space. 
To this end, we examine the eigenspectrum of $G_{\mathbf h}^t$, whose eigenvalues dictate how downstream sensitivity is distributed across activation-space directions, with larger eigenvalues indicating greater sensitivity.
We quantify its concentration using the effective number of directions over which downstream sensitivity is distributed (i.e., participation ratio), defined by $\left(\sum_i \lambda_i\right)^2/\sum_i \lambda_i^2$ where $\lambda_i$ is the i-th eigenvalue.
Across models and layers, we find that capacity is highly concentrated (Figure~\ref{fig:capacity_extra}) with a median effective dimensionality of $24.8$ directions for Gemma, $47.2$ for Llama, $30.1$ for Mistral, and $23.3$ for Qwen, corresponding to just $0.7\%$--$1.2\%$ of the residual stream's dimension. The dimensionality is lowest in early layers ($8-13$) and increases substantially with depth ($37$--$97$).

\paragraph{Localization methods find structures with varying capacity levels}
If downstream sensitivity is concentrated in a small subset of directions, latent structures may differ substantially in how much they overlap with this usable subspace. We therefore compare the median capacity of concept directions identified by DiffMeans and linear probes with that of 16 isotropic random directions.
We normalize all intervention directions to unit norm and sum capacity over a horizon of $n=9$.\footnote{We use 9 offsets, as the causal effect saturates in larger offsets for most concepts (see \S\ref{app:capacity_horizon}).}

We find substantial differences across structure types (Figure~\ref{fig:capacity_extra}). DiffMeans directions have $90.4\%$ more capacity than random directions and $26.4\%$ more than linear probes, while probes have $37.6\%$ more than random directions. This shows that identifiable representational structure can vary substantially in its ability to affect the output, with some structures much more privileged by the model's remaining computation than others.

\textit{Overall, our results suggest that only a low-rank, context-dependent subspace affects the model's output. Thus, identifying usable structure requires distinguishing which parts of the representation lie within this subspace.}

\subsection{Responsiveness: Context Limits Concept Promotion}
\label{sec:responsiveness}
Under softmax, low-probability tokens require larger logit shifts for the same probability increase. Thus, the amount of probability mass assigned to concept-relevant tokens limits the concept's promotability. Factor $\text{\textcolor{cyan}{II}}$, defined as $\sigma_{\mathbf p}(\mathbf r_c)$, reflects this constraint. Since $\mathbf r_c$ indicates concept-relevant tokens, the factor is small when little probability mass is assigned to the concept, but also decreases when the concept is already highly expressed. It is therefore largest in an intermediate regime, where the concept is supported by the current distribution without being saturated. We interpret this factor as the concept's \emph{responsiveness} in the current context.

\paragraph{Experiment}
We test this interpretation by analyzing whether finite, fluency-preserving interventions primarily promote tokens already likely under the unintervened distribution $\mathbf p$, where responsiveness is higher, or ``new'' tokens with little baseline support. We intervene with DiffMeans and random directions at increasing strengths $\alpha$. For each context, we fix the support $\mathcal T$ containing the top $k\in\{1\%,2.5\%,5\%\}$ of tokens under $\mathbf p$, and measure how much of the intervention-induced increase in probability occurs within this support:
\begin{equation}
\phi_{\mathrm{support}}(\alpha)
=
\frac{\sum_{t\in\mathcal T}[p'_t-p_t]_+}
{\sum_{t\in[\mathcal V]}[p'_t-p_t]_+},
\qquad [x]_+ = \max\{x,0\},
\end{equation}
where $\mathbf p'$ is the intervened distribution. The denominator is the total probability mass moved toward higher-probability tokens, while the numerator measures how much of this increase remains within the baseline support.

For each model and layer, we evaluate fluency at 11 intervention strengths $\alpha$ using the same LLM judge as in \citet{wu2025axbench}, which assigns a score of $0,1$ or $2$, with $2$ being fluent. We mark normalized fluency thresholds of $0.8$, $0.7$, and $0.6$, and for each threshold report results up to the largest tested $\alpha$ for which mean fluency remains above it. This allows us to examine how the intervention's effect changes as the model's fluency degrades with increasing intervention strength. Additional details are provided in \S\ref{app:responsiveness_details}.

\begin{figure}[t]
\setlength{\belowcaptionskip}{-10pt}
    \centering
    \includegraphics[width=0.9\linewidth]{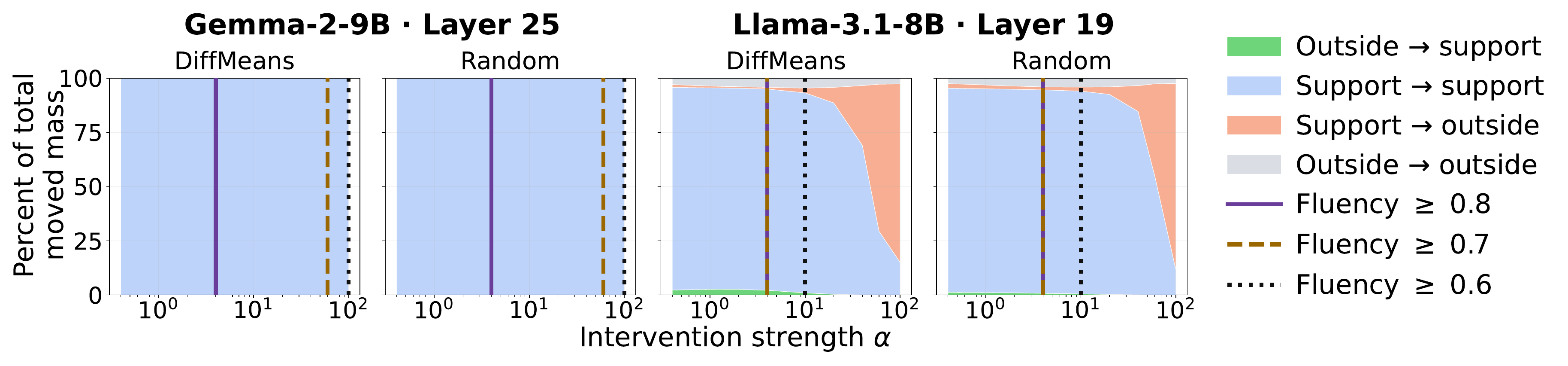}
    \caption{\textbf{Interventions promote outputs already supported by the context.}
    We measure where intervention-induced probability increases are assigned relative to the unintervened distribution. While fluency is preserved, nearly all increased probability goes to tokens in its top $2.5\%$ support.}

    \label{fig:support_flow}
    
\end{figure}

\paragraph{Results}
Figure~\ref{fig:support_flow} shows results for Gemma and Llama for the top $2.5\%$ baseline support at $60\%$ depth (results for additional models, layers and support thresholds show similar trends, see \S\ref{app:responsiveness_full}). At fluency of 0.8, finite interventions (optimized for specific concepts) predominantly redistribute probability among tokens \textit{that were already likely} under $\mathbf p$, with the top $2.5\%$ of baseline tokens receiving approximately $94.5\%$ of the intervention-induced probability increase. Within this fluent regime, the movement occurs predominantly among tokens already in the baseline support (blue regions). Substantial movement outside this support appears only at stronger intervention strengths as fluency degrades ($\alpha \geq 10$, fluency $<0.6$), most clearly for Llama and Mistral, while Gemma and Qwen remain largely confined to the baseline support even at high strengths. The same qualitative behavior holds for DiffMeans and random directions, suggesting that this restriction is primarily imposed by the context rather than by the structure itself. 

\textit{Thus, interventions have limited ability to promote outputs with little support in the current context, showing how the concept's responsiveness constrains the causal influence of a latent structure.}

\subsection{Alignment: Effective Directions Are Context-Dependent}

While movement along a latent structure can affect the model's output, the resulting changes may not promote the target concept. Factor $\text{\textcolor{cyan}{III}}$, defined as $\operatorname{Corr}_{\mathbf p}(\mathbf r_c,J_{\mathbf h}\mathbf v)$ quantifies orientation toward concept-relevant tokens rather than unrelated outputs. 
We therefore view this factor as the structure's \emph{alignment} with the direction that most promotes the concept at the current activation.

Importantly, this interpretation is meaningful only if what constitutes concept-aligned structure depends on context. Otherwise, it would be a measure of how well the structure globally promotes the concept. We therefore test whether the direction that most promotes a concept varies across contexts. To find this direction for an activation $\mathbf{h}$, we estimate the locally-optimal activation-space direction using the closed-form solution to the corresponding regression problem,
\begin{equation}
\mathbf v^*(\mathbf h)
\propto
\left(
G_{\mathbf h}+\lambda I
\right)^{-1}
J_{\mathbf h}^{\top}F_{\mathbf p}\mathbf r_c,
\label{eq:optimal_alignment_direction}
\end{equation}
where $\lambda$ controls the ridge regularization and $G_{\mathbf h}$ is defined in Eq.~\ref{eq:G}.

\begin{figure}[t]
\setlength{\belowcaptionskip}{-8pt}
    \centering
    \includegraphics[width=0.49\linewidth]{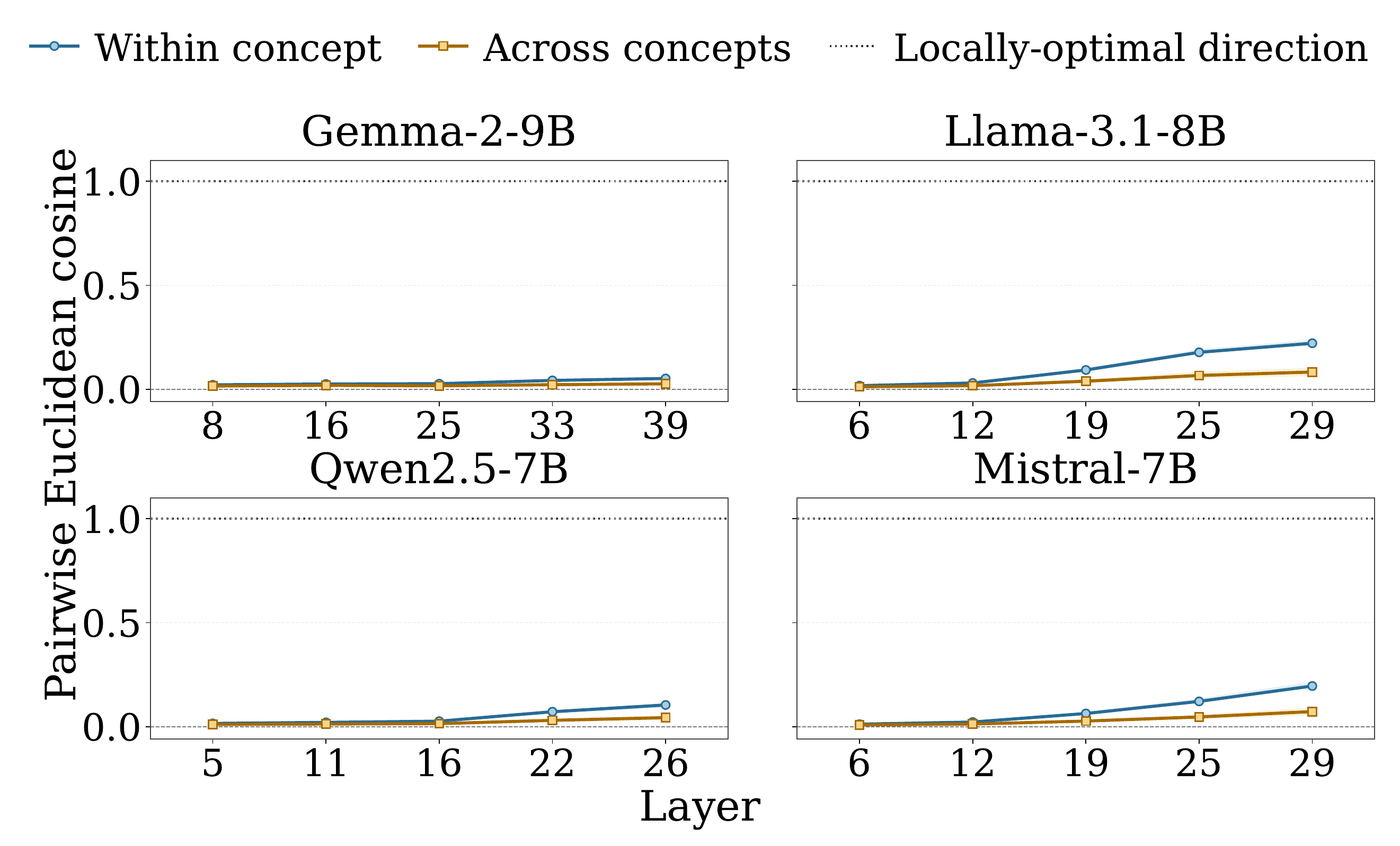}
    \hfill
    \includegraphics[width=0.48\linewidth]{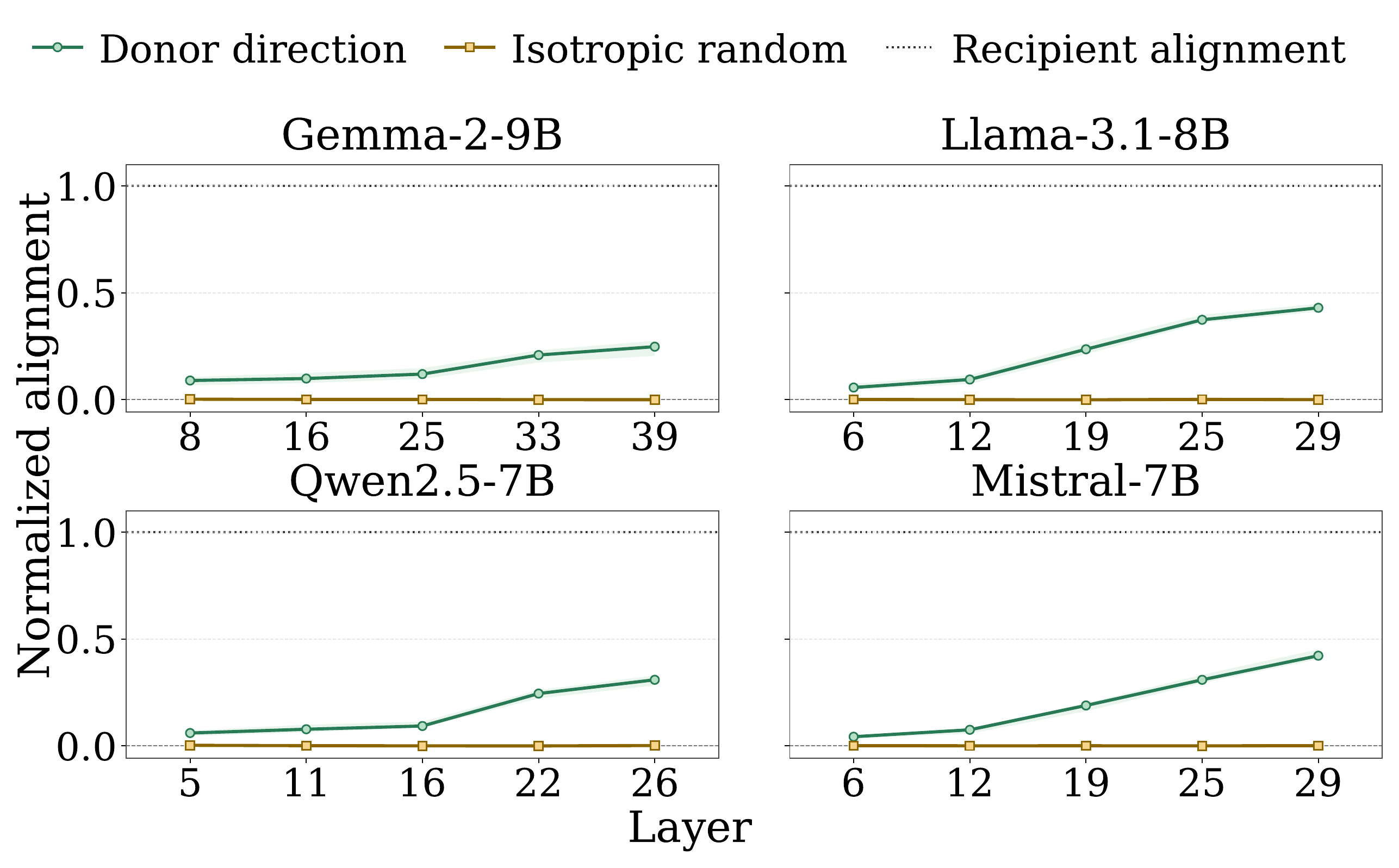}

    \caption{\textbf{Concept-promoting directions are context dependent.}
    We estimate the locally optimal direction for the same concept across contexts.
    \textbf{Left:} Same concept directions are nearly orthogonal in early layers and remain below $0.2$ cosine similarity.
    \textbf{Right:} Same concept directions retain only $10\%-40\%$ of alignment when transferred to other contexts, showing strong context dependence.}

    \label{fig:alignment}
    
\end{figure}

\paragraph{Locally effective directions vary across contexts}
We randomly sample $8$ neutral contexts 
per concept\footnote{We use neutral contexts since in concept-positive contexts even random directions align with the concept due to the constraint described in \S\ref{sec:responsiveness}.} and, at each position and layer, compute the direction in Eq.~\ref{eq:optimal_alignment_direction} independently. Directions are normalized to reflect orientation rather than scale.
We then compare the resulting orientations across activation pairs $(\mathbf h_i, \mathbf h_j)$ using cosine similarity, $\left\langle
\mathbf v^*(\mathbf h_i),
\mathbf v^*(\mathbf h_j)
\right\rangle$.
As a baseline, we also compare directions computed for different concepts.
Figure~\ref{fig:alignment} (left) presents the results.
In early layers, locally-effective directions for the same concept are nearly orthogonal across contexts. Similarity increases in later layers, with directions for the same concept more closely aligned than those for different concepts, however for the majority of models the same concept directions are still distinct with a cosine similarity of ${<}0.2$. This shows that the activation-space direction that most promotes concept-related tokens depends substantially on the current context.

\paragraph{Locally effective directions do not transfer between contexts}
Even if different locally preferred directions are not similar, they could still potentially induce equivalent concept-related effects. We therefore test whether locally effective directions are interchangeable across contexts. 
At each layer we compute the locally preferred direction $\mathbf v_i^*$ independently for each context $i$. We then take the direction $\mathbf v_j^*$ identified in a different donor context $j$ and evaluate it in context $i$, keeping both the context and $\mathbf r_c$ fixed. We measure the transferred direction's alignment in the recipient context $i$, normalized by the alignment of its locally preferred direction, i.e., $\operatorname{Corr}_{\mathbf p_i}
\left(\mathbf r_c, J_i \mathbf v_j^* \right) / 
\operatorname{Corr}_{\mathbf p_i}
\left(\mathbf r_c, J_i \mathbf v_i^* \right)$.

For baselines, we include the normalized alignment of random isotropic directions.
We find that locally effective directions are substantially less aligned when transferred across contexts (Figure~\ref{fig:alignment}, right). In early layers, average normalized alignment falls below 0.1, meaning that a donor direction achieves more than ten times lower alignment in a different context than the recipient's locally preferred direction. While transfer improves in later layers, it remains clearly below the context-local reference, reaching at most $0.4$ normalized alignment.

\textit{Together, our results show that the best direction to promote a concept is strongly context-dependent, causing a fixed latent structure to vary substantially in causal effectiveness across activations.}

\section{Factors Expose Distinct Causal Bottlenecks}
\label{buckets}

Having characterized the factors underlying the first-order causal influence of a structure, we now examine how these factors vary across activations and interact to shape causal influence. First through an illustrative example and then quantitatively.

\begin{figure}[t]
\setlength{\belowcaptionskip}{-8pt}
    \centering
    \begin{subfigure}[t]{0.4\linewidth}
        \centering
    \includegraphics[width=\linewidth]{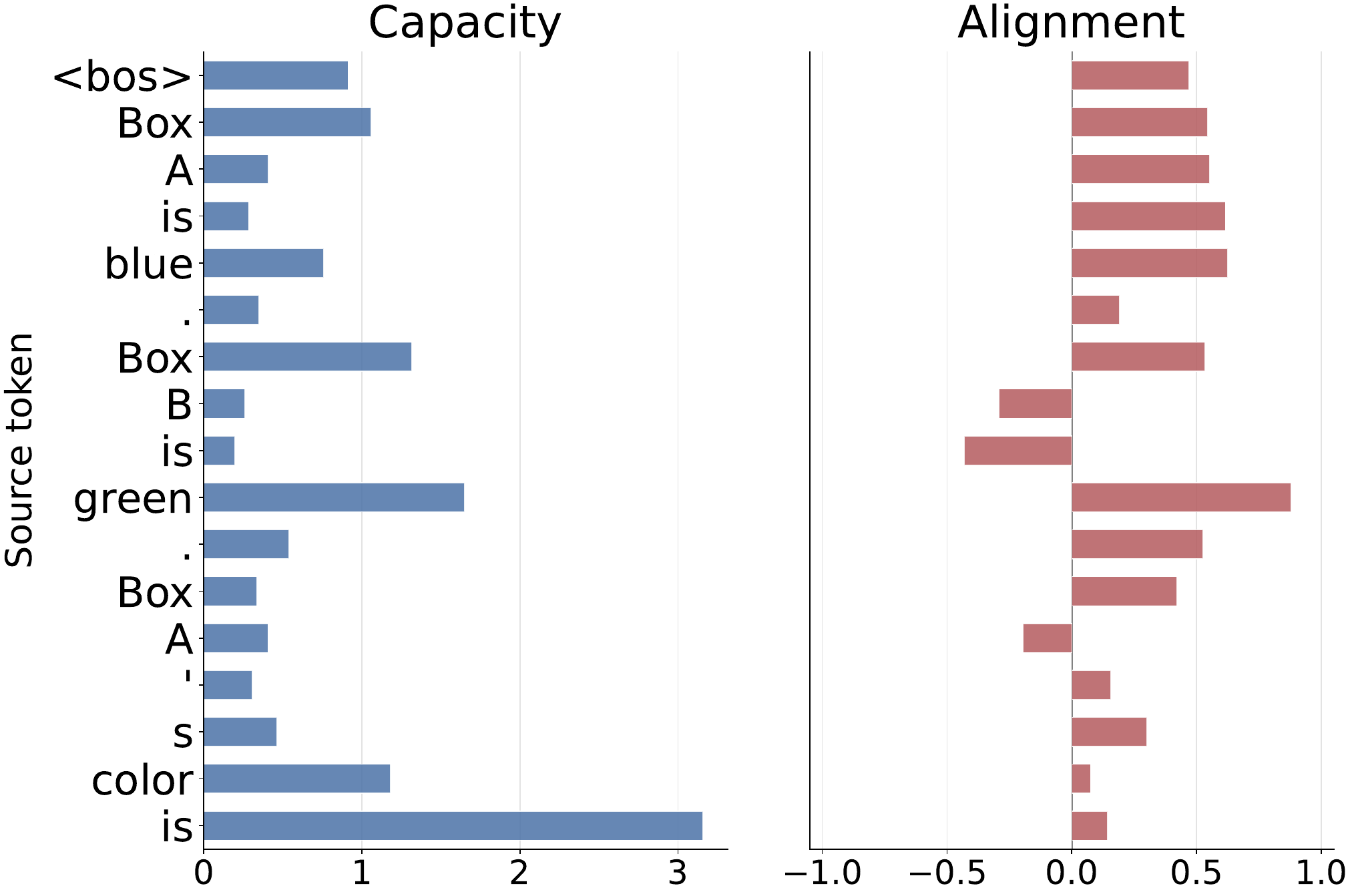}
    \end{subfigure}
    \hfill
    \begin{subfigure}[t]{0.53\linewidth}
        \centering
    \includegraphics[width=\linewidth]{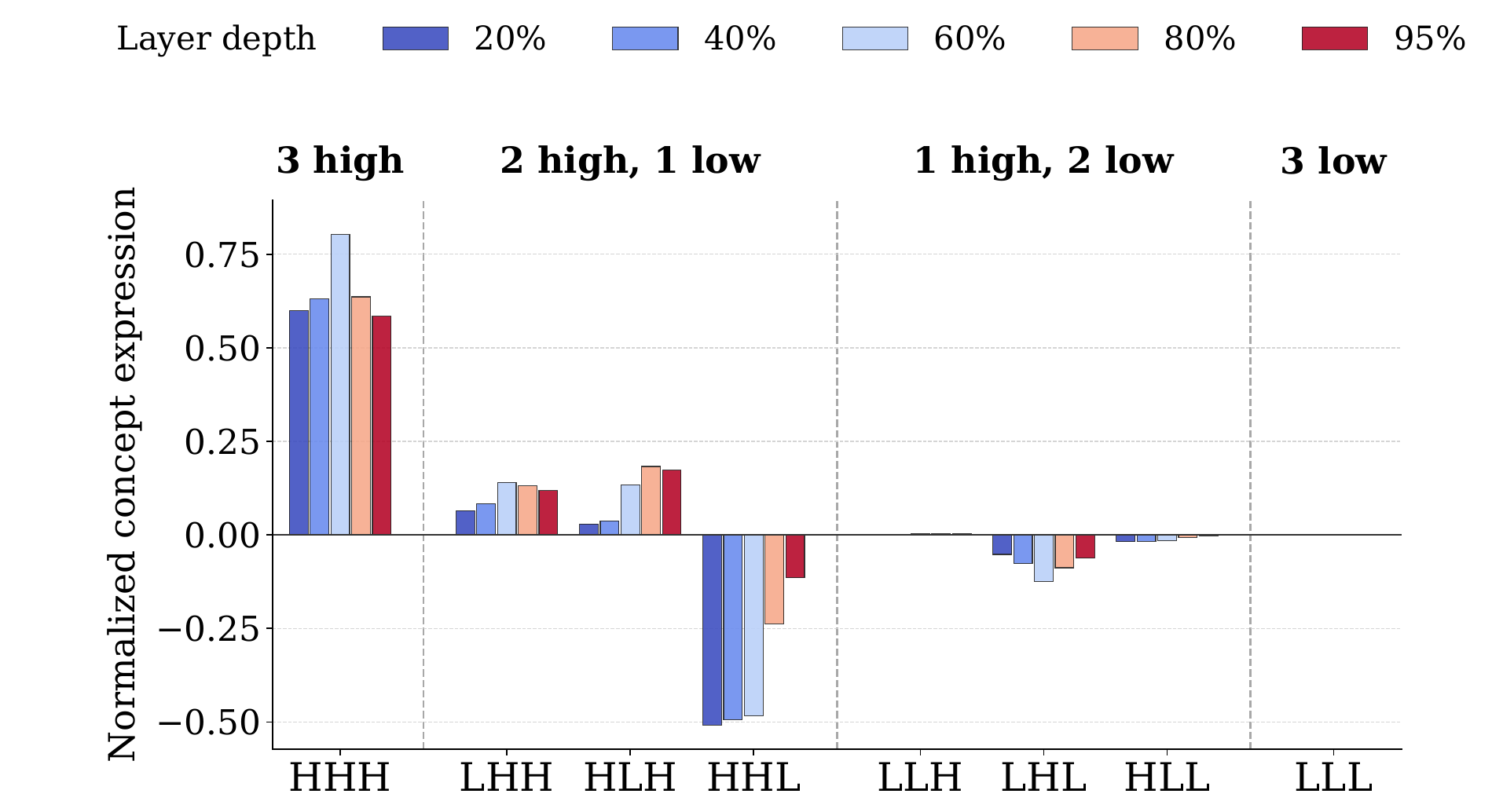}
    \end{subfigure}
    \caption{\textbf{Capacity, responsiveness and alignment impose distinct causal constraints.}
    \textbf{Left:} Capacity and alignment across intervention positions in an example prompt.
    \textbf{Right:} Concept promotion across all eight high/low factor combinations; a low value of any factor sharply reduces the effect, while low alignment suppresses the concept.}
    \label{fig:buckets}
\end{figure}

\paragraph{Example} Consider a simple binding task, where the model must associate given entities with their corresponding attributes: \textit{``Box A is blue. Box B is green. Box A's color is''}. We apply a DiffMeans direction for \emph{green} at each source position and measure its effect on the probability of \emph{green} at the final position. Because the target position is fixed, responsiveness is constant, allowing us to isolate capacity and alignment. Figure~\ref{fig:buckets} shows the capacity and alignment associated with intervening at each source position. Although the final position has the highest capacity, more than $3\times$ of earlier positions, its alignment with the \emph{green} direction is much lower than at the earlier ``green'' token ($0.14$ vs.\ $0.88$). Consequently, intervening at ``green,'' despite occurring $7$ positions earlier, has a $3.2\times$ larger first-order effect on the probability of the target token \emph{green}.

\paragraph{Experiment}
We sample $16,000$ contexts from AllenAI C4 \citep{raffel2020exploring}, each consisting of $n=256$ tokens, and for each context measure capacity, responsiveness and alignment with respect to each of the 50 concept covectors $\mathbf r_c^T$ computed in \S\ref{experimental_setting}. Then, we intervene with the corresponding DiffMeans direction $\mathbf v_c$ at a source position $s \in \{n, n-1, ...,n-7\}$ and measure the resulting change in concept expression at the target position $t=n$, where $s=t$ captures the effect at the intervention position itself. Given the intervened and unintervened output distributions $\mathbf p_t'$ and $\mathbf p_t$ at position $t$, we measure concept expression as $
\mathbf r_c^\top
\left(
\mathbf p_t'-\mathbf p_t
\right)
$.
Holding the model, layer, and concept fixed, we treat each evaluated source--target pair as one sample. Each sample is assigned a high or low label per factor, using the top and bottom quartiles of that factor across samples. Quartile thresholds are estimated with five-fold cross-fitting over contexts. The three binary labels partition the retained samples into eight joint factor settings.
We then compare the intervention effect across these groups.

\paragraph{Results}
We find that the three factors impose distinct constraints, with all eight high/low combinations represented in the data (full bucket counts in \S\ref{app:bucket_counts}). Figure~\ref{fig:buckets} shows concept expression for each combination aggregated across models, revealing that a single unfavorable factor is sufficient to sharply reduce intervention efficacy. Low capacity and low responsiveness reduce concept expression by $84\%$ and $95\%$, respectively, while low alignment can reverse the effect and suppress the concept. When all three factors are low, concept promotion is effectively zero.  Per-model plots showing similar trends are provided in \S\ref{app:bucket_per_model}.
In \S\ref{app:effective_positions}, we further validate the need for all three factors to be high over longer generations by showing that their product identifies positions where steering is especially effective relative to low-product or randomly selected positions.

\section{Accounting for Causal Constraints Improves Steering}

In this section, we demonstrate the utility of our findings for effective steering. We constrain linear probes to the high-capacity subspaces identified by our analysis and show that this substantially improves steering while largely preserving concept detection.

\paragraph{Causal probes} For each concept $c$ and layer, we identify the activation-space directions with the largest downstream capacity. Specifically, for each sample in the positive and neutral datasets, we take the activation $\mathbf h$ at source position $s$, compute $G_{\mathbf h}^{t}$ (Eq.~\ref{eq:G}) with respect to each target position $t=s,\ldots,s+7$ and average these matrices across samples and source--target pairs. The leading $r$ eigenvectors of this average define a high-capacity subspace $Q_{c}$ of rank $r$. We then train a linear probe as described in \S\ref{experimental_setting}, while constraining it to lie in $Q_{c}$ by projecting it back onto the subspace after each gradient step. We refer to the resulting classifier as a \emph{causal probe}.

\paragraph{Evaluation}
For each model, layer, concept, and structure, we steer generations from 16 fixed neutral prefixes by adding $\alpha\mathbf v$ at every decoding step. We sweep 14 strengths $\alpha\in[0.4,100]$, generate up to 50 tokens, and score concept expression and fluency similarly to \citet{wu2025axbench}, reporting their harmonic mean.\footnote{In \citet{wu2025axbench} they use a scale of 0-2, we report normalized scores.} We select the best $\alpha$ for each setting using five-fold cross-validation over prefixes and report scores across all 50 concepts. We compare causal probes with $r\in\{8,16,32,64,128\}$ against standard probes, DiffMeans, and a strong supervised baseline ReFT-r1 \cite{wu2025axbench}. See \S\ref{app:steering_evaluation} for full details.

\begin{figure}[t]
\setlength{\belowcaptionskip}{-10pt}
    \centering
    \includegraphics[width=0.99\linewidth]{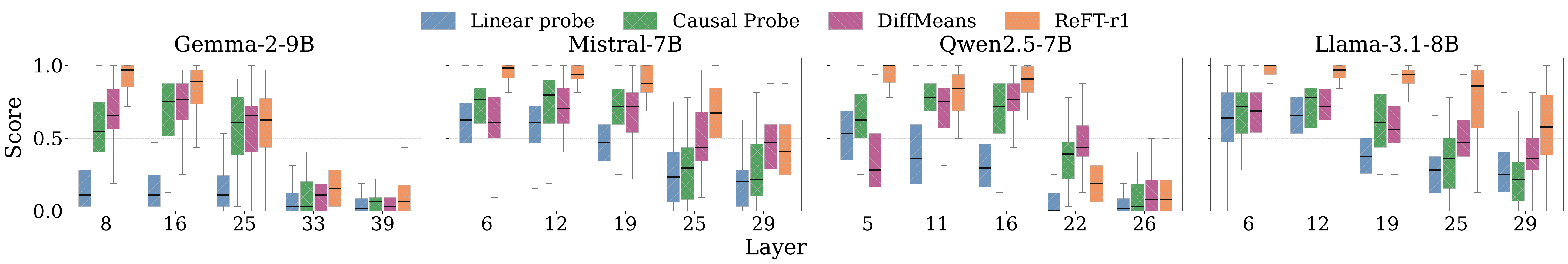}
    \caption{\textbf{Causal probes improve steering.}
    We construct causal probes by restricting linear probes to high-capacity subspaces and evaluate their steering performance.
    Causal probes substantially outperform standard probes, often performing similarly to DiffMeans.}
    \label{fig:causal_probes}
    
\end{figure}

\paragraph{Results}
Final scores are shown for $r=128$ in Figure~\ref{fig:causal_probes}, with additional ranks and concept and fluency scores in \S\ref{app:steering_components}. Causal probes substantially improve steering over raw probes, particularly in early and middle layers. Across models, the median improvement is $118\%$ for Gemma, $57\%$ for Qwen, $23\%$ for Mistral, and $17\%$ for Llama. Notably, these gains come with only a small decrease in concept detection, with a median classification accuracy of $90.4\%$, compared to $93.8\%$ for the unconstrained probe. These results show that learned probes capture predictive structure along directions with little causal capacity, and removing these causally ineffective directions substantially improves steering while largely preserving concept detection.

\section{Related Work}

\paragraph{The detection--causality gap}
Prior work has identified the disparity between structure in activation space and causality. Early work showed that probing accuracy can reflect the capacity of the probe itself, and that representations may encode properties that are not relevant to the current task of the model \citep{hewitt2019designing, ravichander-etal-2021-probing}. \citet{elazar2021amnesic} made this distinction causal, showing that probing accuracy need not track the behavioral importance of the detected information. Recent work further extends our understanding of this gap, showing that where information is localized is not necessarily where editing is most effective \citep{hase2023does}, and structures that accurately detect a concept often produce weak or unreliable effects when used for steering \citep{arad2025saes, wu2025axbench, galeone2026perfectdetectionfailedcontrol}. \citet{hoang2026sparseautoencodersencodeconcepts} also showed that SAE features can have different downstream effects across contexts. While previous work identified a gap between representation and control, we quantify this gap and decompose it into concrete, interpretable factors, showing that causal influence is fundamentally context dependent.

\paragraph{Jacobians and output space geometry}
Recent work utilized output-space geometry to understand the activation-space of LMs. \citet{park2026information} studied the information geometry induced by softmax and uses it to derive geometry-aware probing and steering. \citet{wurgaft2026manifold} pulled back the output-space geometry with the model's Jacobian to construct nonlinear steering paths. \citet{wang2026fishbackpullbackfishergeometry} likewise pulled the Fisher geometry of the output distribution into activation space and used the resulting metric to derive improved steering directions. These works use output-space geometry to design or optimize interventions, instead we ask why the causal influence of a given latent structure varies across contexts, and which distinct factors govern this variation.

Recently, \citet{hernandez2024linearity, gurnee2026verbalizablerepresentationsformglobal,blank2026rlens} utilized Jacobians to connect activation-space directions to future outputs. These methods identify a useful subspace by aggregating Jacobian effects across contexts and identifying directions with consistent downstream effects. In contrast, we study why the causal efficacy of a given latent structure varies across contexts, using the local Jacobian and output distribution to characterize the factors governing its effect.

\section{Conclusion}

By measuring concept expression at the output and pulling it back to intermediate activations, we cast the causal influence of a latent structure as a product of three factors: \emph{capacity}, \emph{responsiveness}, and \emph{alignment}. Across four models, we find that these factors capture distinct, interpretable constraints: downstream sensitivity is concentrated in a small, context-dependent subset of directions, interventions primarily promote outputs already supported by the context, and concept-promoting directions vary substantially across contexts. These constraints explain how a structure can be strongly represented yet remain weak or unreliable for control. We further show that removing predictive but causally ineffective directions with \emph{Causal Probes} improves median steering by 29\%, with only a 3\% drop in concept-detection accuracy. Overall, our results show that causal effectiveness is not intrinsic to a latent structure, but depends on its relationship with the current context. Effective localization and control therefore require accounting not only for \emph{what} is represented, but also \emph{when and how} that representation can be causally utilized.

\subsubsection*{Acknowledgments}
This work was supported in part by a grant from Coefficient
Giving, the Academic Research Program at Google, and the
Israel Science Foundation grant 1083/24.

\bibliography{iclr2027_conference}
\bibliographystyle{iclr2027_conference}

\appendix

\section{Experimental Details}
\label{app:experimental_details}

\subsection{Dataset Construction and Prompts}
\label{app:data}

We sampled 50 concepts from the list utilzied in \citet{wu2025axbench} and paired them
with 304 concept-independent prefixes. For each concept and prefix, we generated
a concept-positive completion, along with one neutral completion per prefix,
using gpt-4.1-mini \citep{openai2024gpt4ocard}. This produced 15,200 positive and 304 neutral
examples. We split the data by prefix to prevent overlap across the training,
validation, and test sets. The complete dataset is included in the provided
repository, and the generation prompts are reproduced below.

We then partitioned the 304 prefixes into training, validation, and test sets using
a random 80/10/10 split. All positive and neutral examples
sharing a prefix were assigned to the same split, preventing the same context
from appearing in both training and evaluation. This produced 243 training,
31 validation, and 30 test prefixes.

Example prefixes include ``The main takeaway is'', ``The important question is'', ``A more general way to say this is'', ``The idea is useful because'', ``The answer depends on'', ``The simplest case is when'', ``The main source of confusion is'', and ``A useful comparison is''.

\begin{tcolorbox}[
  title=Concept-Positive Completion Prompt,
  colback=white,
  colframe=black,
  boxrule=0.5pt,
  arc=2pt,
  left=6pt,right=6pt,top=6pt,bottom=6pt
]
\ttfamily\small
\raggedright
You are generating a dataset example for studying language model activations.

Concept description:

\{concept\_description\}

Prefix:

\{prefix\}

Task:

Complete the prefix so that the resulting text naturally and clearly expresses the concept.
Meaning the completion should exemplify the concept very clearly.

Rules:

- Do not change the prefix.

- Continue directly from the prefix.

- The continuation should be natural, fluent, and specific.

- The concept should be clearly present, but do not sound like you are defining the concept.

- Do not mention "concept", "dataset", "positive sample", or this instruction.

- Avoid repetitive phrasing.

- Write one sentence only.

- Return only the continuation, not the full text or anything else.
\end{tcolorbox}

\begin{tcolorbox}[
  title=Neutral Completion Prompt,
  colback=white,
  colframe=black,
  boxrule=0.5pt,
  arc=2pt,
  left=6pt,right=6pt,top=6pt,bottom=6pt
]
\ttfamily\small
\raggedright
You are completing a sentence.

Prefix:

\{prefix\}

Task:

Complete the prefix with a normal continuation one could expect to see.

Rules:

- Do not change the prefix.

- Continue directly from the prefix.

- Write one sentence only.

- Return only the continuation, not the full text or anything else.
\end{tcolorbox}

\subsection{Concept Covector Construction}
\label{app:covectors}

For each concept $c$, we collected the tokens that appeared in its positive
training completions but never appeared in the neutral completions. We built
these sets separately for each model tokenizer and excluded special tokens,
such as beginning-of-sequence (BOS), end-of-sequence (EOS), padding, and
unknown tokens. For each token $y$ in
the model vocabulary, let $n_c^{+}(y)$ denote its number of occurrences in
positive completions for concept $c$, and let $n^{0}(y)$ denote its number of
occurrences in neutral completions. We define
\begin{equation}
\mathcal V_c=\{y:n_c^{+}(y)>0,\;n^{0}(y)=0\},
\qquad
\mathbf r_c(y)=
\begin{cases}
1, & y\in\mathcal V_c,\\
0, & y\notin\mathcal V_c.
\end{cases}
\end{equation}

Thus, $\mathbf r_c$ assigns weight one to tokens associated with concept $c$
and zero to all other tokens, thus capturing the verbalized aspect of a concept.

\subsection{Latent Structure Construction}
\label{app:latent_structures}

We trained one direction for each concept, model, and layer using only the
prefix-grouped training split. For every positive and neutral completion, we
collected the residual-stream activation after the selected transformer block
at each position predicting a completion token, using all 243 positive and
243 neutral training examples per concept.

For DiffMeans, we averaged the positive and neutral activations separately and
used their difference:
\begin{equation}
\mathbf v_c^{\mathrm{DM}}
=
\frac{\boldsymbol{\mu}_c^{+}-\boldsymbol{\mu}^{0}}
{\left\|\boldsymbol{\mu}_c^{+}-\boldsymbol{\mu}^{0}\right\|_2},
\end{equation}
where $\boldsymbol{\mu}_c^{+}$ and $\boldsymbol{\mu}^{0}$ are the mean positive
and neutral activations, respectively.

The linear probe was a bias-free logistic classifier trained on the same
activations using binary cross-entropy. Probe weights were initialized from
$\mathcal N(0,0.01^2)$. The final probe
weight was normalized to unit Euclidean norm before being used as a direction.
The constrained probe variants used the exact same setting,
but restricted the learned direction to the corresponding rank-128 causal
subspace.

For ReFT-r1, we followed the rank-one TopKReLU setting from \citet{wu2025axbench}.

\subsection{Jacobian and Capacity Computation}
\label{app:jacobian_computation}

For an activation $\mathbf h$ and output position $t$, let
$J_{\mathbf h}^t$ denote the Jacobian of the output logits with respect to
$\mathbf h$. For future output positions, the tokens between the source and
target positions are fixed to the observed continuation using teacher forcing.

Because the full Jacobian has shape $|\mathcal V|\times d$ it is inefficient to
construct explicitly, we compute only its products with vectors using
automatic differentiation. A Jacobian--vector product (JVP) gives the
first-order change in the logits caused by moving $\mathbf h$ along a direction
$\mathbf v$. We then apply the Jacobian of softmax at the original output
distribution, converting this logit change into a first-order probability
change.

A vector--Jacobian product (VJP), implemented by backpropagation, then maps
this change back to activation space. This computes
\[
G_{\mathbf h}^t\mathbf v
=(J_{\mathbf h}^t)^\top F_{\mathbf p^t}J_{\mathbf h}^t\mathbf v
\]
without explicitly constructing either $J_{\mathbf h}^t$ or
$G_{\mathbf h}^t$. Capacity is the probability-weighted standard deviation of
the logit changes produced by the JVP. All derivatives are computed using
automatic differentiation.

\section{Additional Capacity Analysis}
\label{app:capacity}

\subsection{Derivation of the Local KL Approximation}
\label{app:kl_derivation}

For a small intervention $\alpha\mathbf v$, the logits at position $t$ satisfy
\begin{equation}
\Delta \mathbf z
=
\alpha J_{\mathbf h}^t\mathbf v
+
O(\alpha^2).
\end{equation}
For a softmax distribution, the KL divergence between the original and perturbed
distributions has the second-order expansion
\begin{equation}
D_{\mathrm{KL}}
\left(
\mathbf p
\,\Vert\,
\mathbf p_{\Delta\mathbf z}
\right)
=
\frac{1}{2}
\Delta\mathbf z^\top
F_{\mathbf p}
\Delta\mathbf z
+
O(\|\Delta\mathbf z\|^3),
\end{equation}
where
$
F_{\mathbf p}
=
\operatorname{diag}(\mathbf p)-\mathbf p\mathbf p^\top
$
is the softmax Fisher matrix. Substituting
$\Delta\mathbf z=\alpha J_{\mathbf h}^t\mathbf v$ gives
\begin{equation}
D_{\mathrm{KL}}
\left(
\mathbf p
\,\Vert\,
\mathbf p_{\alpha\mathbf v}
\right)
=
\frac{\alpha^2}{2}
\mathbf v^\top
(J_{\mathbf h}^t)^\top
F_{\mathbf p}
J_{\mathbf h}^t
\mathbf v
+
O(\alpha^3)
=
\frac{\alpha^2}{2}
\mathbf v^\top
G_{\mathbf h}^t
\mathbf v
+
O(\alpha^3).
\end{equation}
Thus, for sufficiently small interventions,
\begin{equation}
D_{\mathrm{KL}}
\left(
\mathbf p
\,\Vert\,
\mathbf p_{\alpha\mathbf v}
\right)
\approx
\frac{\alpha^2}{2}
\mathbf v^\top
G_{\mathbf h}^t
\mathbf v.
\end{equation}

\subsection{Capacity Prediction Across Future Positions}
\label{app:capacity_offsets}

\begin{figure}[t]
    \centering
    \includegraphics[width=\linewidth]{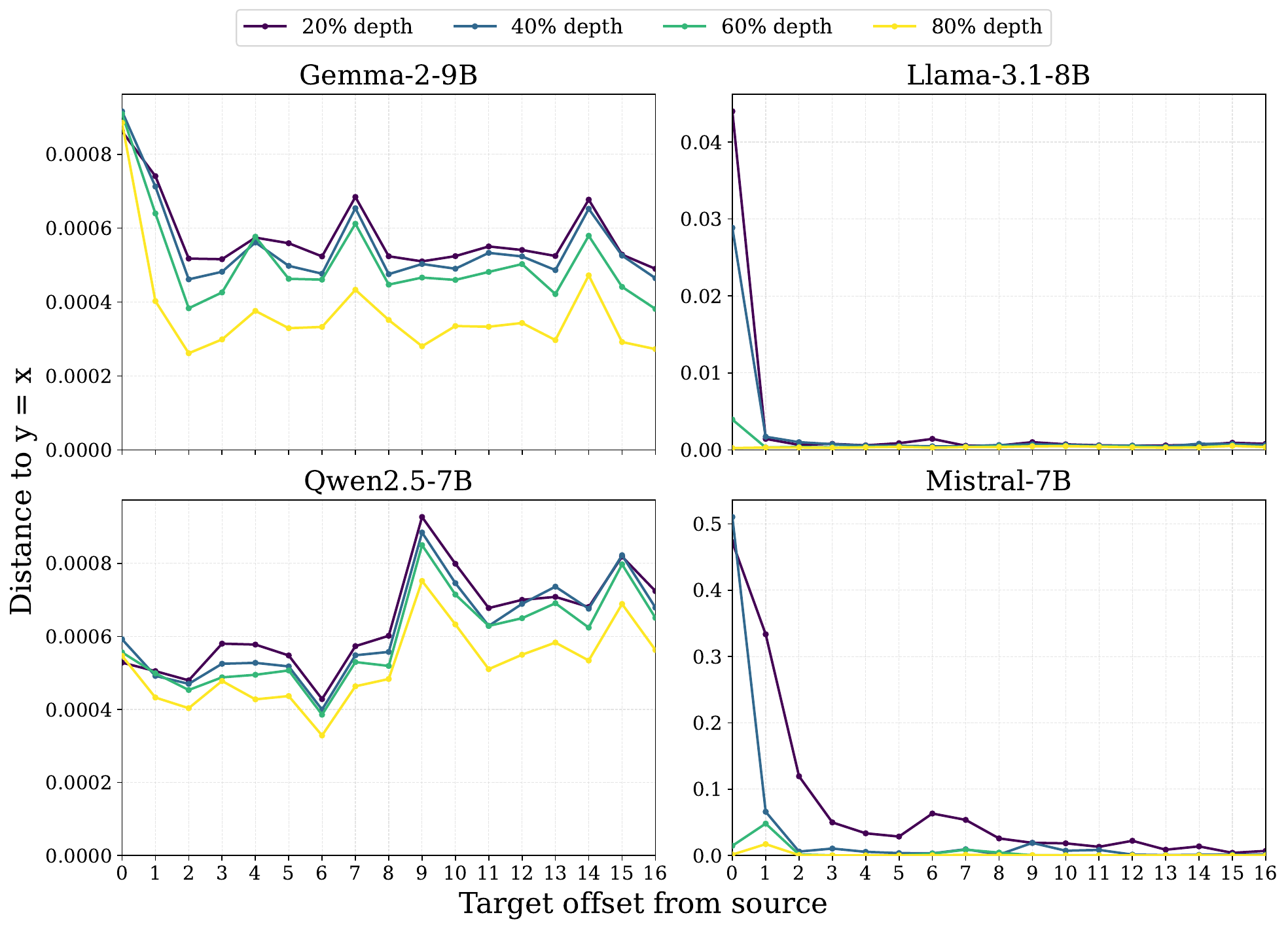}
    \caption{\textbf{Predicted KL closely tracks actual KL across future offsets and layers.} We report approximation error as the distance from \(y=x\) between predicted and actual KL for offsets \(t=0,\ldots,16\). Trends are consistent across layers, with error generally decreasing at later offsets, except for a slight increase in Qwen.
}
    \label{fig:additional_offsets_capacity}
\end{figure}

We include in Figure~\ref{fig:additional_offsets_capacity} additional results for offset $t$ up to $t=16$, for the analysis done in \S\ref{sec:capacity_interp}. We report the results as the distance from $y=x$ for predicted and actual KL. The larger the value, the bigger the approximation error. The figure shows that across layers the trends are very similar. Looking across offsets, the distance decreases across models, with Qwen being the exception with a slight increase. Generally, we see predicted KL tracks actual KL closely.

\subsection{Choice of Sequence Horizon}
\label{app:capacity_horizon}

\begin{figure}[t]
    \centering
    \includegraphics[width=\linewidth]{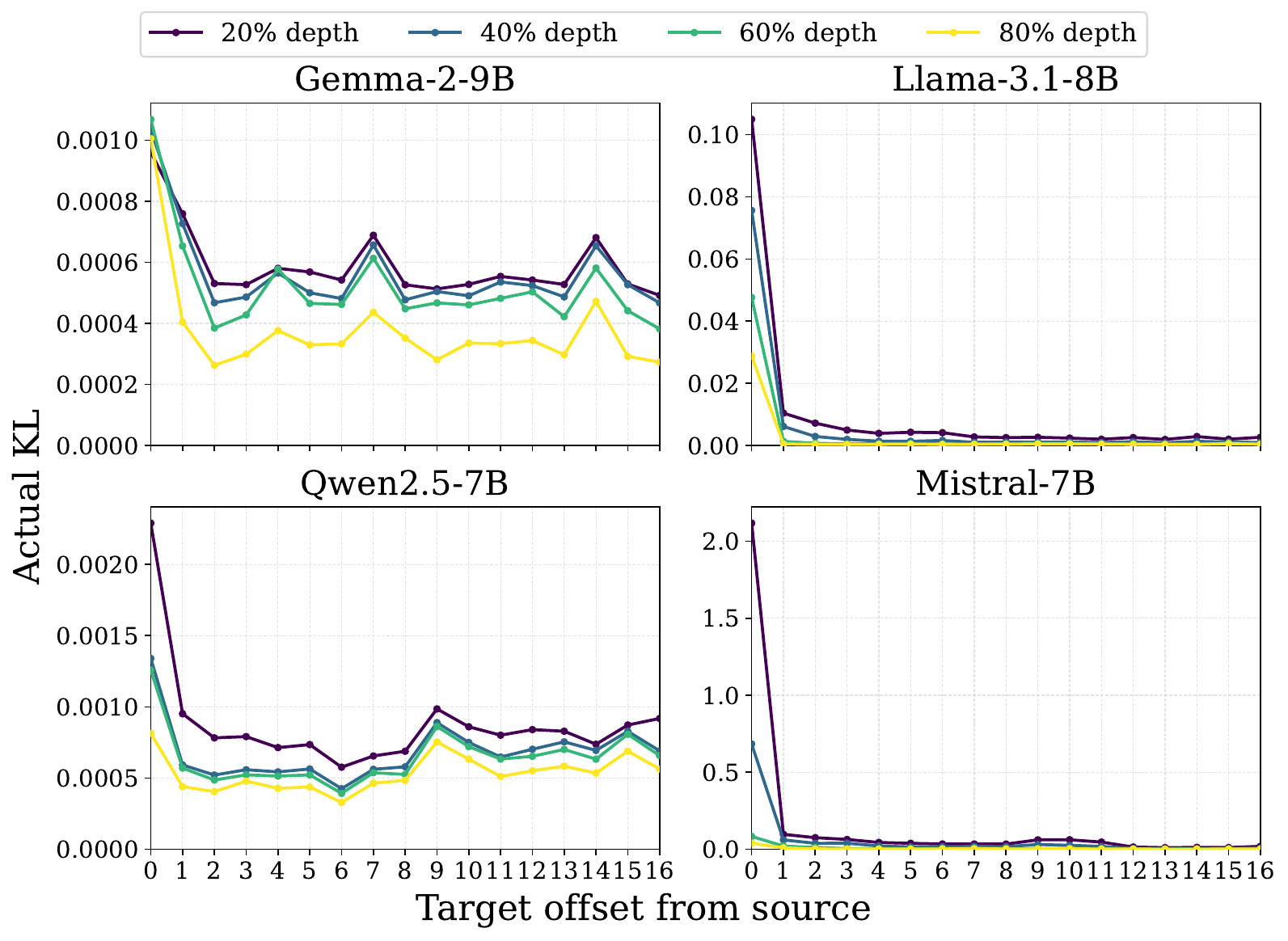}
    \caption{Actual KL induced by an intervention decreases with offsets, consistently across models and layers.
    }
    \label{fig:additional_offsets_capacity_kl}
\end{figure}

Figure~\ref{fig:additional_offsets_capacity_kl} shows the actual KL produced by interventions across offsets. We can see that the KL on future offsets decreases significantly with the offset position. This is true for all models, concepts and layers tested. Therefore, adding additional offsets to the analysis does not affect the results and trends observed. As such in this analysis we decided to include up to 8 additional offsets beyond the intervention position.

\section{Additional Responsiveness Analysis}
\label{app:responsiveness}

\subsection{Experimental and Fluency-Evaluation Details}
\label{app:responsiveness_details}

We evaluate intervention strengths $\alpha\in\{0.4,0.8,1.2,1.6,2,4,10,20,40,60,100\}$. For each model, layer, and concept, the random baseline is sampled from an isotropic Gaussian in activation space and normalized to unit Euclidean norm. Fluency is evaluated by Gemini-2.5-Flash-Lite \citep{comanici2025gemini25pushingfrontier} using the fluency prompt from \citet{wu2025axbench}. Scores are then normalized. For each model, layer, and strength, we compute the unweighted mean fluency across all judged DiffMeans generations, pooling concepts and contexts. For each target $\tau\in\{0.8,0.7,0.6\}$, the fluency boundary is the largest tested $\alpha$ whose mean score is at least $\tau$.

\subsection{Results Across Layers and Support Thresholds}
\label{app:responsiveness_full}

Figure~\ref{fig:support_flow_all} provides the results for all models,
4 layers, and both random and DiffMeans interventions, and support sizes: the top $1\%$, $2.5\%$,
and $5\%$ of tokens. The figure shows that across all layers and models, the trend is the same. In the fluent regime any intervention, whether learned or random, promotes the support which captures concept responsive tokens. This captures the context dependence of causality. If a concept is not responsive in the current context, then intervening even with a well localized structure, will not produce the intended effect while remaining fluent.

\begin{figure*}[p]
    \centering
    \includegraphics[width=\linewidth]
    {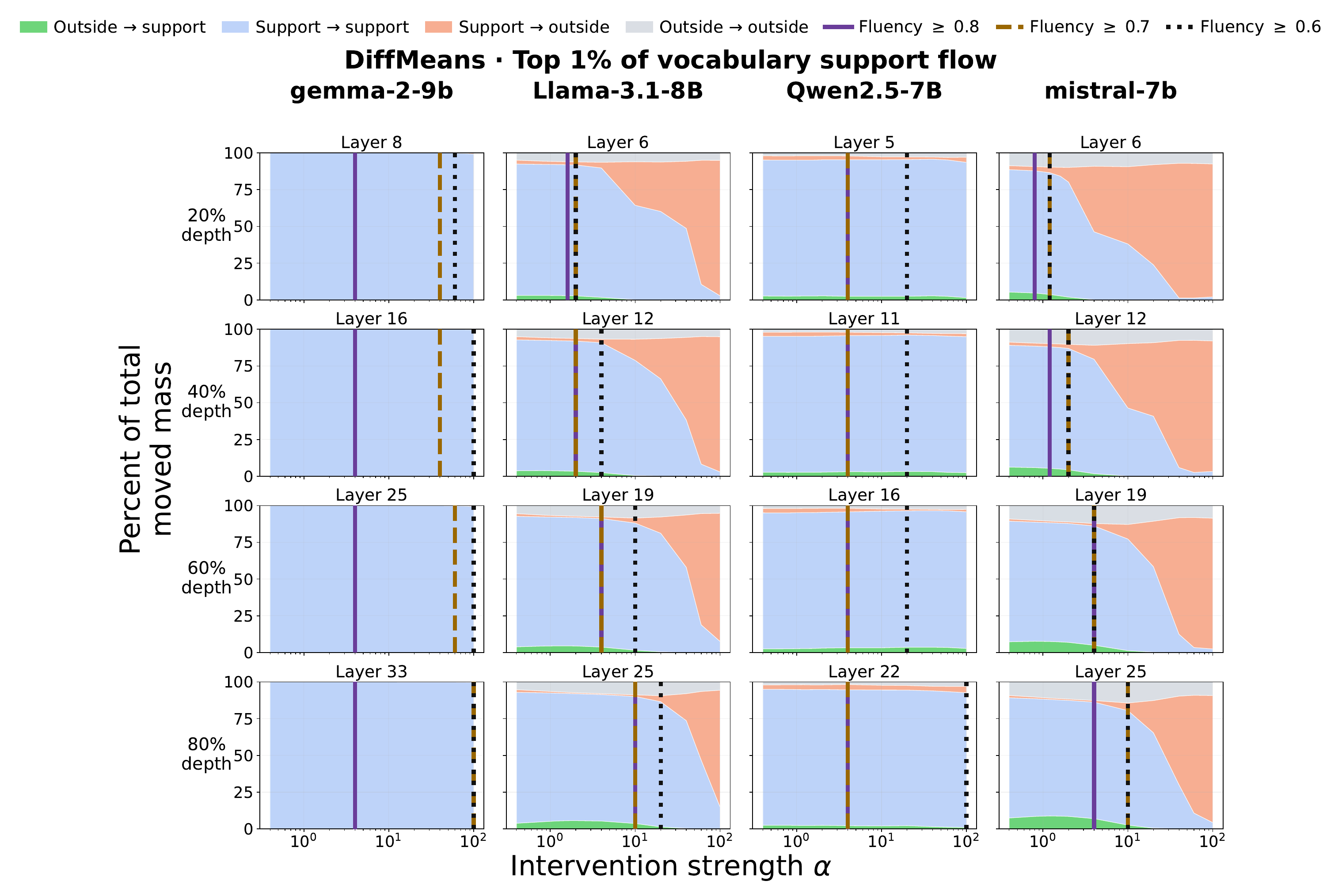}
    \caption{
    \textbf{Interventions primarily promote outputs already supported by the context.}
    Support-flow results across models, layers, intervention directions, and support sizes.
    In the fluent regime, most intervention-induced probability increase remains within the
    baseline support, consistently across settings.
    }
    \label{fig:support_flow_all}
\end{figure*}

\begin{figure*}[p]
    \ContinuedFloat
    \centering
    \includegraphics[width=\linewidth]
    {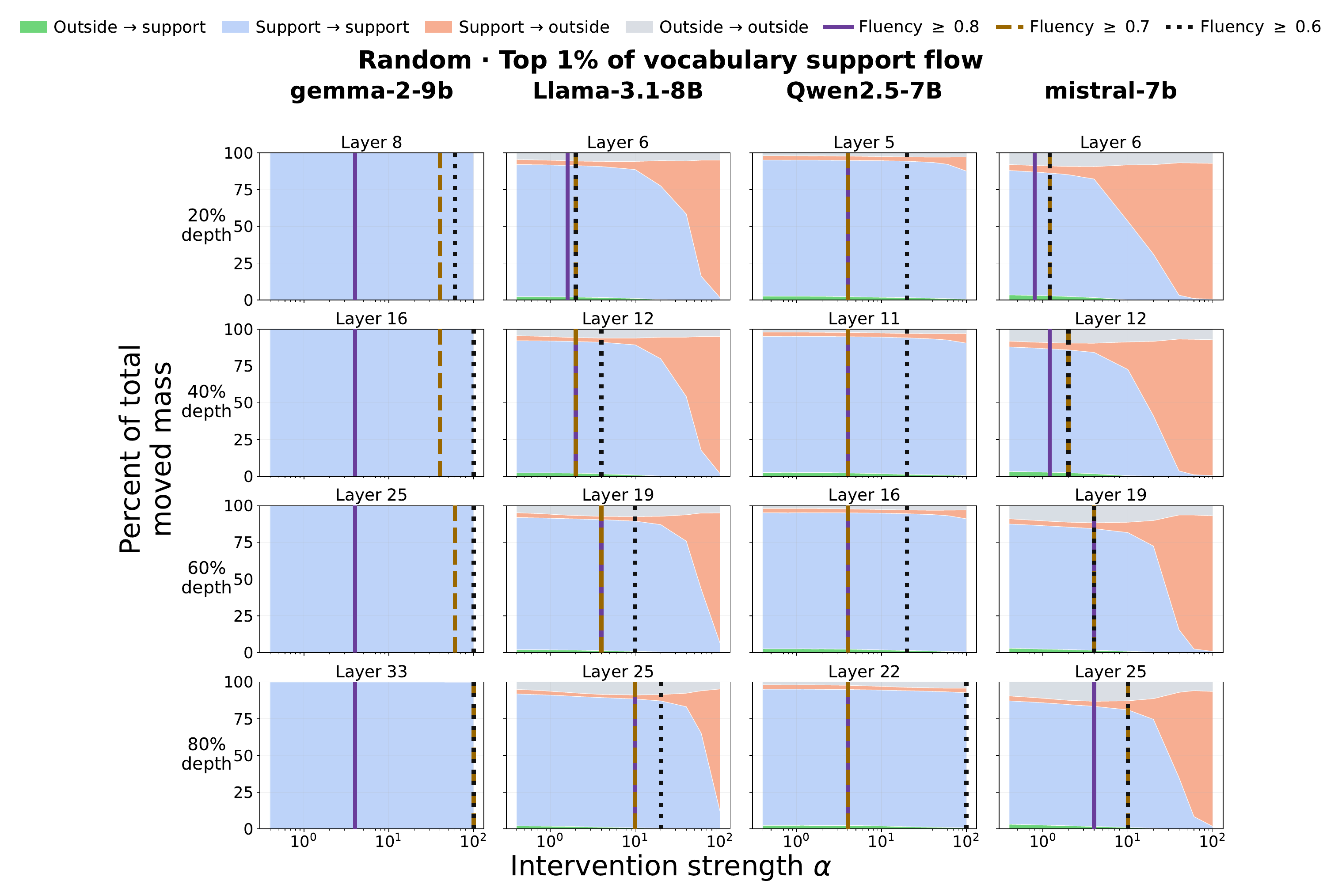}
    \caption[]{\textbf{Interventions primarily promote outputs already supported by the context.} (continued)}
\end{figure*}

\begin{figure*}[p]
    \ContinuedFloat
    \centering
    \includegraphics[width=\linewidth]
    {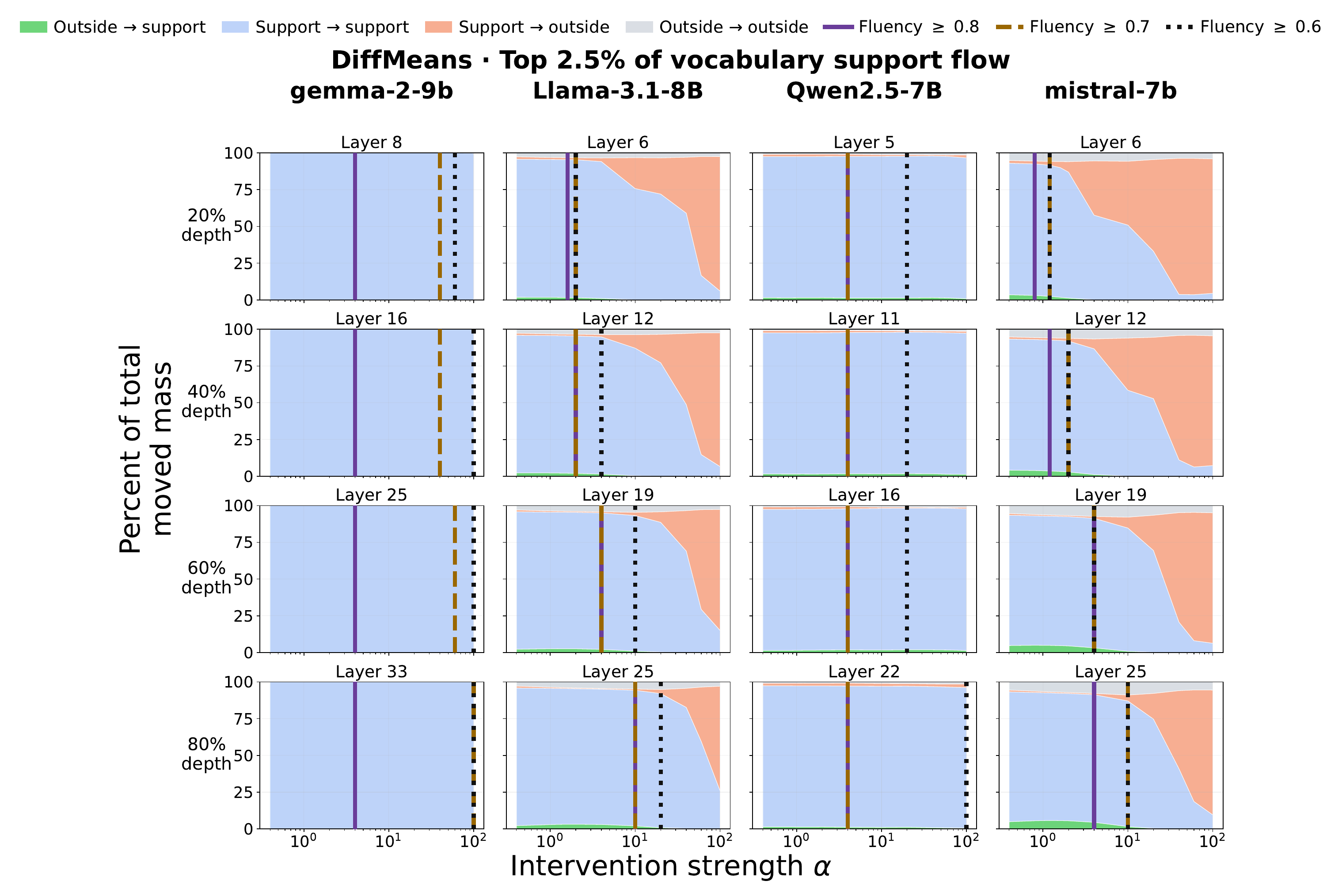}
    \caption[]{\textbf{Interventions primarily promote outputs already supported by the context.} (continued)}
\end{figure*}

\begin{figure*}[p]
    \ContinuedFloat
    \centering
    \includegraphics[width=\linewidth]
    {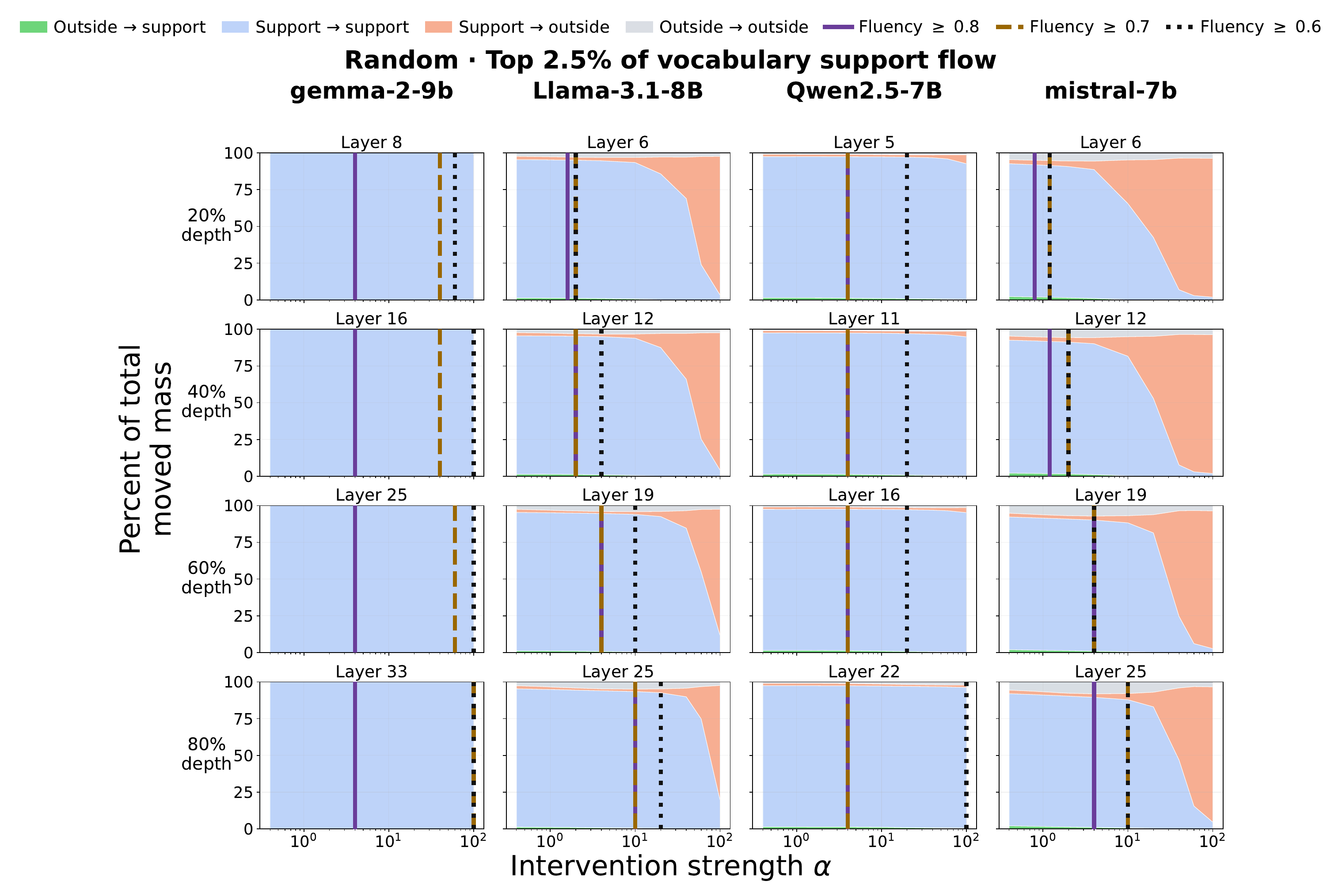}
    \caption[]{\textbf{Interventions primarily promote outputs already supported by the context.} (continued)}
\end{figure*}

\begin{figure*}[p]
    \ContinuedFloat
    \centering
    \includegraphics[width=\linewidth]
    {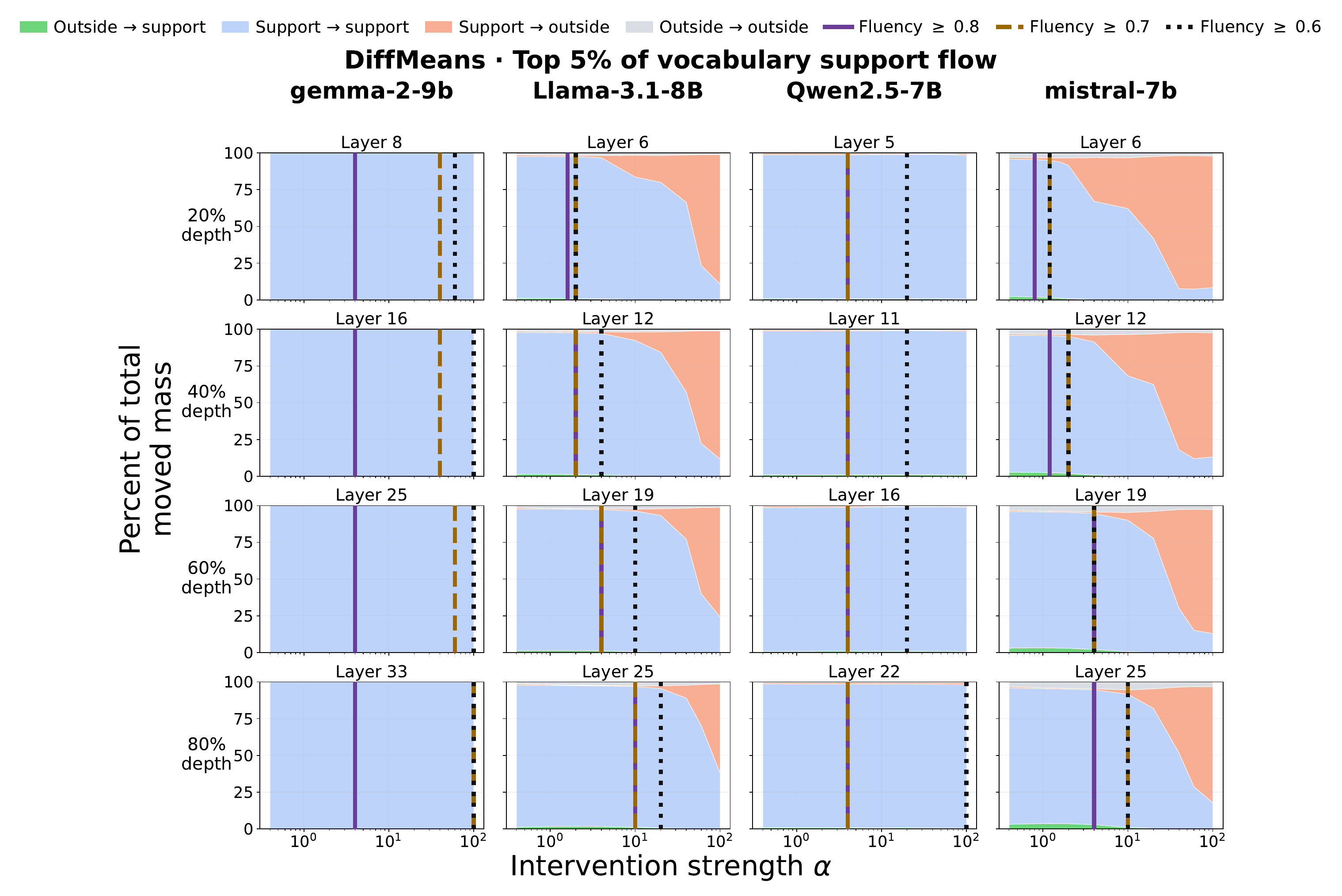}
    \caption[]{\textbf{Interventions primarily promote outputs already supported by the context.} (continued)}
\end{figure*}

\begin{figure*}[p]
    \ContinuedFloat
    \centering
    \includegraphics[width=\linewidth]
    {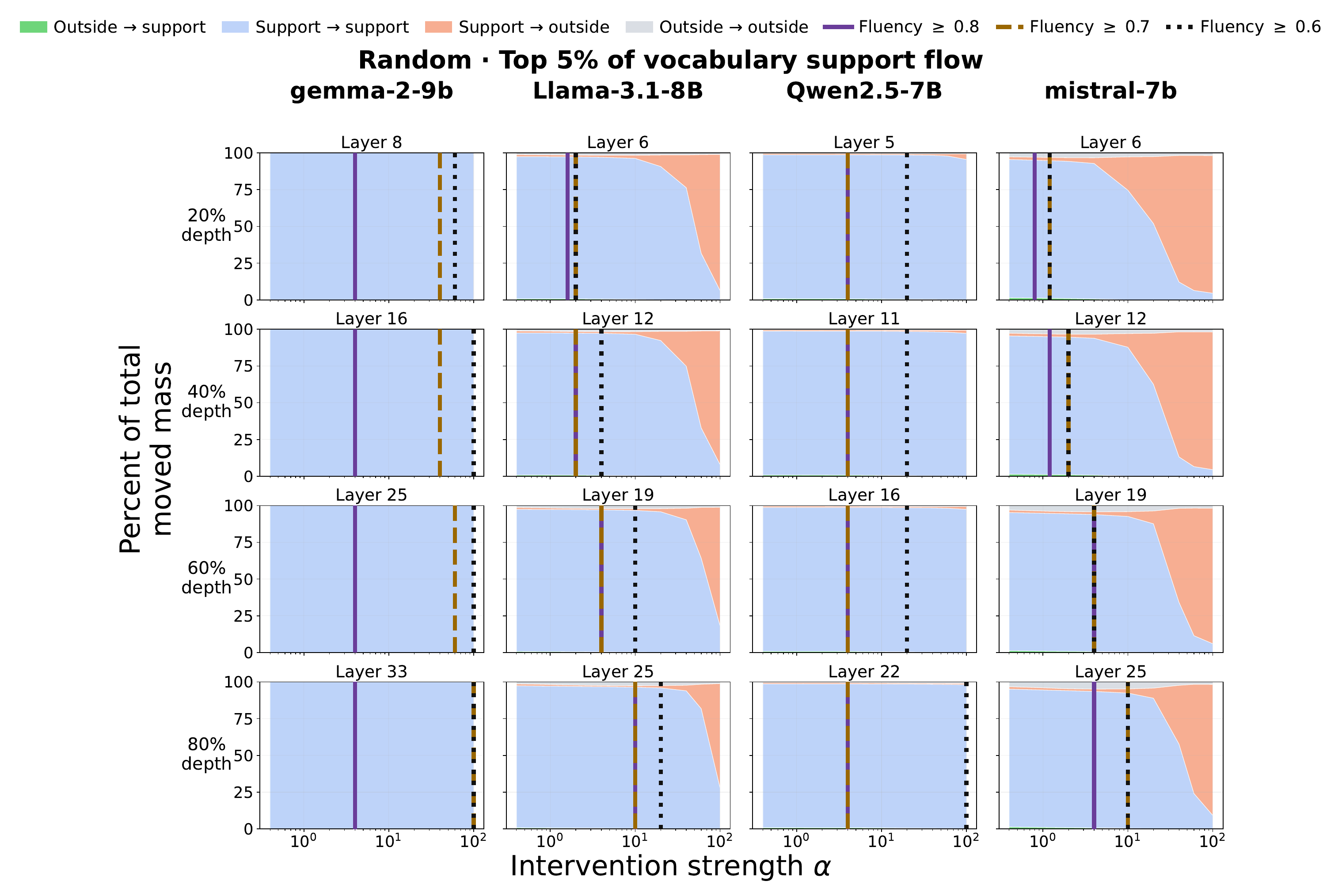}
    \caption[]{\textbf{Interventions primarily promote outputs already supported by the context.} (continued)}
\end{figure*}

\section{Additional Alignment Details}
\label{app:alignment}

\subsection{Locally Optimal Direction Estimation}
\label{app:alignment_details}

For each concept and context, we solve for the activation direction that most
efficiently promotes the concept. We obtain it by approximately solving
$(G_i+10^{-4}I)\tilde{\mathbf v}_i=g_i$. We add the small ridge term to make the
solution numerically stable. Because $G_i$ is too large to construct (scales with vocabulary size), we use
automatic differentiation and conjugate gradient for at most 32 iterations.
To check the approximation, we recompute the true solution error every four
iterations, retain the iterate with the smallest error, and verify that a
direction transferred to its own context gives the expected normalized score
of one (within $0.002$). The direction is then scaled to have unit capacity.
We use the same eight held-out neutral texts for every concept, sampling up to
16 token positions from each.

\section{Additional Analysis of the Three Causal Factors}
\label{app:factors}

\subsection{Full Factor-Bucket Statistics}
\label{app:bucket_counts}

\begin{figure*}[t]
    \centering
    \includegraphics[width=\linewidth]{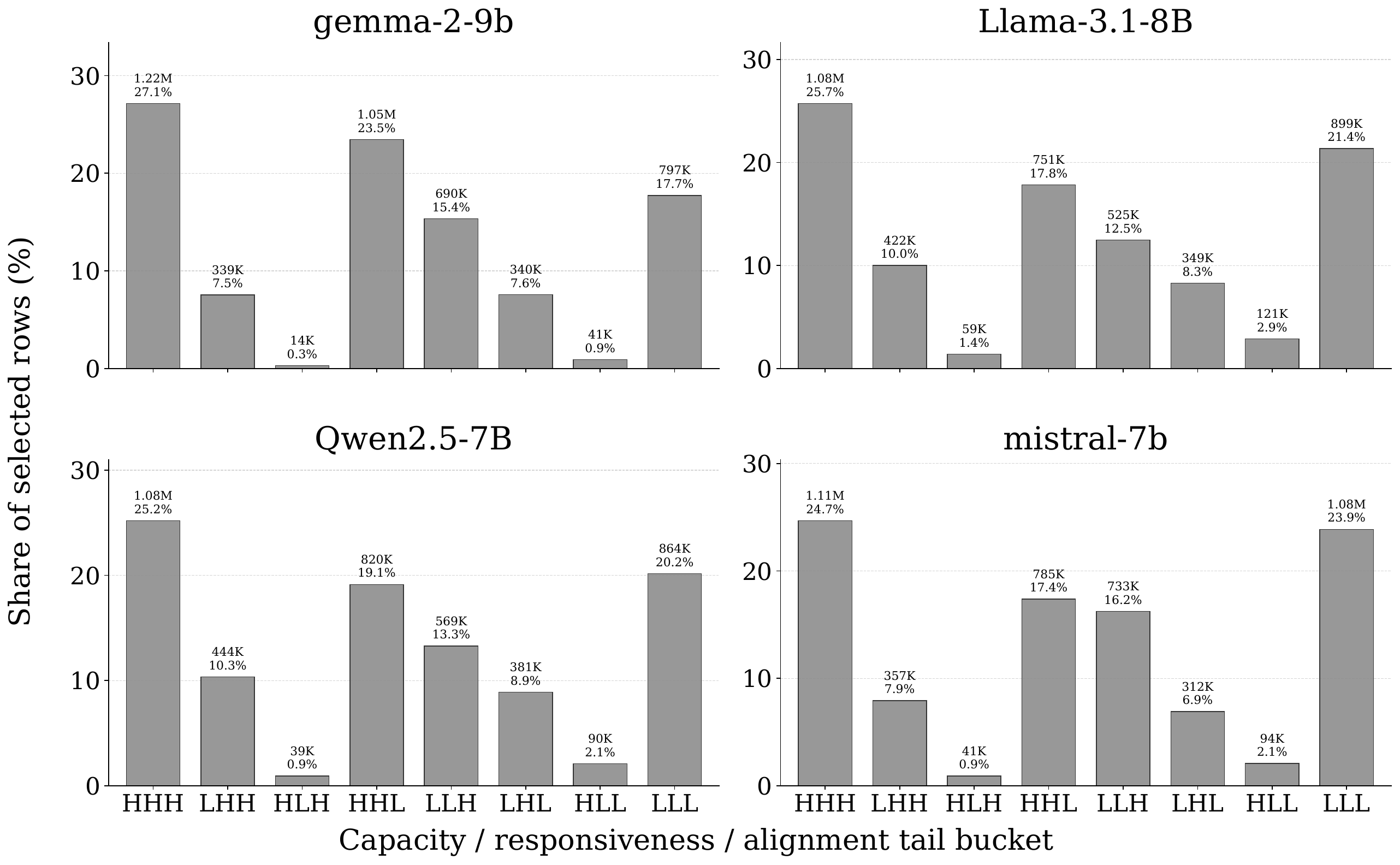}
    \caption{
    \textbf{Distribution of capacity--responsiveness--alignment factor combinations.}
    The fraction of retained samples in each high/low combination for each model.
    }
    \label{fig:bucket_distribution}
\end{figure*}

Figure~\ref{fig:bucket_distribution} shows the distribution of the eight factor combinations presented in \S\ref{buckets}. We find that cases in which exactly one factor is low are common, comprising roughly $26\%{-}31\%$ of retained samples across models. As shown in \S\ref{buckets}, a low value of any single factor can substantially reduce causal effectiveness, making these cases particularly important. We additionally find that there is an uncommon regime when responsiveness is low and capacity is high.

\subsection{All Model Factor-Bucket Plots}
\label{app:bucket_per_model}

\begin{figure*}[p]
    \centering
    \includegraphics[width=\linewidth]
    {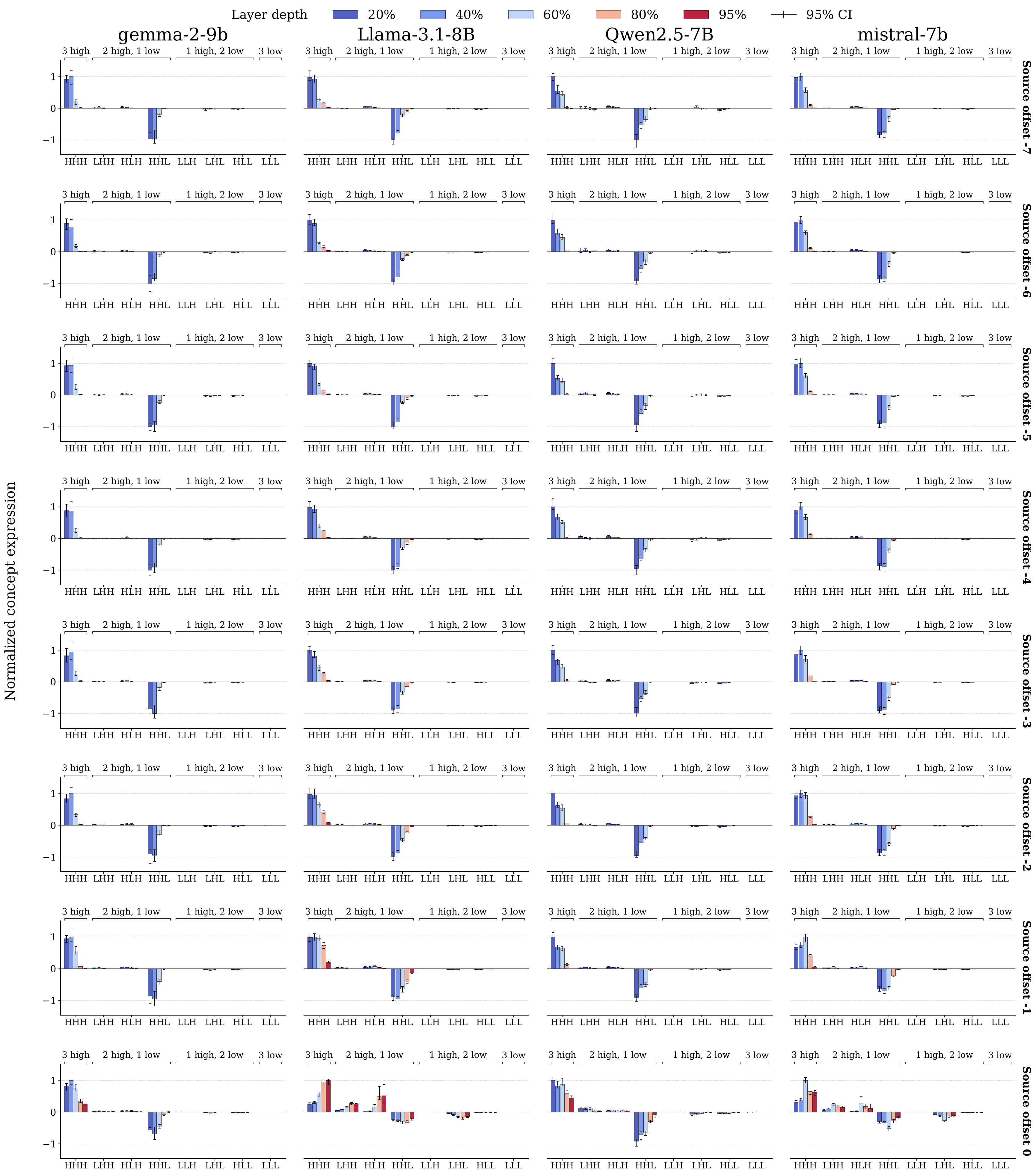}
    \caption{
    \textbf{Factor-bucket results across source offsets.}
    Normalized concept promotion for each responsiveness--capacity--alignment
    combination across models, layers, and source offsets $-7$ through $0$.
    The same qualitative pattern persists across offsets: promotion is strongest
    when all three factors are high, while a low value of any factor substantially
    limits the effect.
    }
    \label{fig:bucket_offsets}
\end{figure*}

Figure~\ref{fig:bucket_offsets} extends the factor-bucket analysis across offsets $0$ through $-7$. 
The same qualitative pattern persists across offsets: concept promotion is strongest when all three factors are high, while low capacity or responsiveness sharply reduces the effect and low alignment can reverse it, making a localized structure suppress the concept it was intended to promote.

\subsection{Steering on Effective Positions}
\label{app:effective_positions}

Our decomposition shows that the causal effectiveness of a fixed direction
varies across positions according to its responsiveness, capacity, and
alignment. We therefore ask whether these factors can identify \emph{where}
steering should be applied. At decoding step $t$, for each preceding
position $i$, we score the effect of intervening along concept direction
$\mathbf v_c$ at position $i$ on the next-token distribution $\mathbf p_t$
as
\begin{equation}
R_i^t
=
\mathbf r_c^\top
F_{\mathbf p_t}
J_i^t
\mathbf v_c,
\qquad
J_i^t
=
\frac{\partial \mathbf z_t}{\partial \mathbf h_i}.
\end{equation}
Thus, $R_i^t$ is the signed first-order effect of steering position $i$ on
concept expression at the current next-token prediction, and corresponds to
the product of the three factors in
Eq.~\ref{eq:causal-effect-decomposition}.

\paragraph{Experiment}
We use the same steering and evaluation setup as above, but intervene on
only a subset of positions. At each decoding step $t$, we compute $R_i^t$
for all $n$ active positions and select exactly
$k=\lfloor n/4\rfloor$ positions. We compare steering performance at the
positions with the highest $R_i^t$, at those with the lowest $R_i^t$, and
at an equal number selected uniformly at random. We add
$\alpha\mathbf v_c$ only at the selected positions, sample the next token,
and recompute the selection at the following decoding step.

\begin{table*}[t]
\caption{\textbf{Steering at positions with high predicted causal effect
improves performance.}
Concept, fluency, and final steering scores when intervening at randomly
selected positions, positions with the highest $R_i^t$, or positions with
the lowest $R_i^t$. Values report mean $\pm$ standard deviation across
concepts.}
\label{tab:steering_realizability_avg}
\centering
\resizebox{\textwidth}{!}{%
\begin{tabular}{@{}lccc ccc ccc ccc@{}}
\toprule
& \multicolumn{3}{c}{\textbf{Qwen2.5-7B}}
& \multicolumn{3}{c}{\textbf{Gemma-2-9B}}
& \multicolumn{3}{c}{\textbf{Llama-3.1-8B}}
& \multicolumn{3}{c}{\textbf{Mistral-7B}} \\
\cmidrule(lr){2-4}
\cmidrule(lr){5-7}
\cmidrule(lr){8-10}
\cmidrule(l){11-13}

\textbf{Metric}
& Rand. & High $R$ & Low $R$
& Rand. & High $R$ & Low $R$
& Rand. & High $R$ & Low $R$
& Rand. & High $R$ & Low $R$ \\
\midrule

Final
& $0.22 \pm .14$ & $\mathbf{0.29 \pm .15}$ & $0.18 \pm .14$
& $0.15 \pm .15$ & $\mathbf{0.27 \pm .16}$ & $0.15 \pm .17$
& $0.30 \pm .14$ & $\mathbf{0.36 \pm .15}$ & $0.23 \pm .14$
& $0.25 \pm .13$ & $\mathbf{0.40 \pm .14}$ & $0.22 \pm .15$ \\

Concept
& $0.26 \pm .14$ & $\mathbf{0.33 \pm .15}$ & $0.25 \pm .13$
& $0.13 \pm .15$ & $\mathbf{0.26 \pm .15}$ & $0.14 \pm .16$
& $0.37 \pm .14$ & $\mathbf{0.47 \pm .14}$ & $0.31 \pm .14$
& $0.36 \pm .12$ & $\mathbf{0.49 \pm .14}$ & $0.33 \pm .14$ \\

Fluency
& $0.73 \pm .09$ & $\mathbf{0.77 \pm .08}$ & $0.69 \pm .12$
& $\mathbf{0.91 \pm .02}$ & $0.88 \pm .04$ & $0.90 \pm .03$
& $\mathbf{0.68 \pm .09}$ & $0.67 \pm .13$ & $0.61 \pm .14$
& $0.62 \pm .16$ & $\mathbf{0.72 \pm .10}$ & $0.59 \pm .17$ \\

\bottomrule
\end{tabular}%
}
\end{table*}

\paragraph{Results}
Table~\ref{tab:steering_realizability_avg} summarizes the results.
Steering high-$R_i^t$ positions improves the final score over the random
baseline in 19 of 20 settings, by $44.2\%$ on average, whereas steering
low-$R_i^t$ positions reduces performance by $15.5\%$ and underperforms
random in $80\%$ of settings. Thus, high values identify positions where
steering is particularly effective, while low values identify positions
where causal constraints actively limit the intervention. These results
show that the factors remain predictive over longer generations and provide
a useful model of where steering will be effective.

\section{Additional Steering Details and Results}
\label{app:steering}

\subsection{Causal Probe Training Details}
\label{app:causal_probe_details}

For each concept and layer, we average the capacity matrices (Eq.~\ref{eq:G}) over all completion positions in the positive and neutral training examples, using the current output and up to seven subsequent outputs. Each available source--target pair receives equal weight. Because the matrices are too large to construct explicitly, we apply them using automatic differentiation and estimate their leading eigenvectors with randomized subspace iteration. We check this approximation using eigenvector residuals, orthogonality, and comparisons with full model computations. The first $r\in\{8,16,32,64,128\}$ eigenvectors define the rank-$r$ capacity subspace. 

Within each subspace, we train a bias-free logistic probe, projecting its gradient and weight back into the subspace after every update. Training uses AdamW for 30 epochs, with learning rate $10^{-3}$, batch size 1024, and the final direction is normalized to unit length.

\subsection{Steering Evaluation and Judge Prompts}
\label{app:steering_evaluation}

For each concept and intervention method, we generated 50 tokens from the same 16 fixed neutral prefixes using temperature $0.8$ and nucleus sampling with $p=0.95$. The intervention was applied at every decoding step. We evaluated $ \alpha\in \{0.4,0.8,1.2,1.6,2,3,4,6,8,10,20,40,60,100\}.$ To select $\alpha$ without evaluating and selecting it on the same prefixes, we divided the 16 prefixes deterministically into five folds of sizes $4,3,3,3,3$. For each model, layer, concept, and method, one fold was held out, and the strength with the highest mean harmonic score on the other four folds was selected. The selected strength was then evaluated only on the held-out fold. Repeating this procedure for all five folds produced one out-of-fold score for each of the 16 prefixes, which were then averaged for each concept.

The concept-expression and fluency prompts are provided. We applied the judging separately for each completed sentence, including both the prefix and generated continuation. Judging used \texttt{gemini-2.5-flash-lite} with thinking disabled. Each judge returned a score in $\{0,1,2\}$, which we divided by two. The final score was the harmonic mean of normalized concept expression and fluency, set to zero if either score was zero.

\begin{tcolorbox}[
  title=Concept Expression Judge Prompt,
  colback=white,
  colframe=black,
  boxrule=0.5pt,
  arc=2pt,
  left=6pt,right=6pt,top=6pt,bottom=6pt
]
\ttfamily\small
\raggedright
Act as an impartial scientific evaluator.

Evaluate whether the specified concept is expressed in the completed sentence.

Ignore grammar when deciding concept expression.

Score 0 if absent, 1 if present weakly or awkwardly, and 2 if clearly and naturally expressed.

Return only JSON with integer field "score" and short string field "rationale".

Concept:

\{concept\}

Completed sentence:

\{sentence\}
\end{tcolorbox}

\begin{tcolorbox}[
  title=Fluency Judge Prompt,
  colback=white,
  colframe=black,
  boxrule=0.5pt,
  arc=2pt,
  left=6pt,right=6pt,top=6pt,bottom=6pt
]
\ttfamily\small
\raggedright
Act as an impartial scientific evaluator.

Evaluate only the fluency of the completed sentence, ignoring concept relevance.

Score 0 if highly unnatural or incomprehensible, 1 if understandable with noticeable errors, and 2 if fluent.

Return only JSON with integer field "score" and short string field "rationale".

Completed sentence:

\{sentence\}
\end{tcolorbox}

\subsection{Concept and Fluency Scores}
\label{app:steering_components}

\begin{figure*}[t]
    \centering
    
    \begin{subfigure}{0.72\linewidth}
        \centering
        \includegraphics[width=\linewidth]{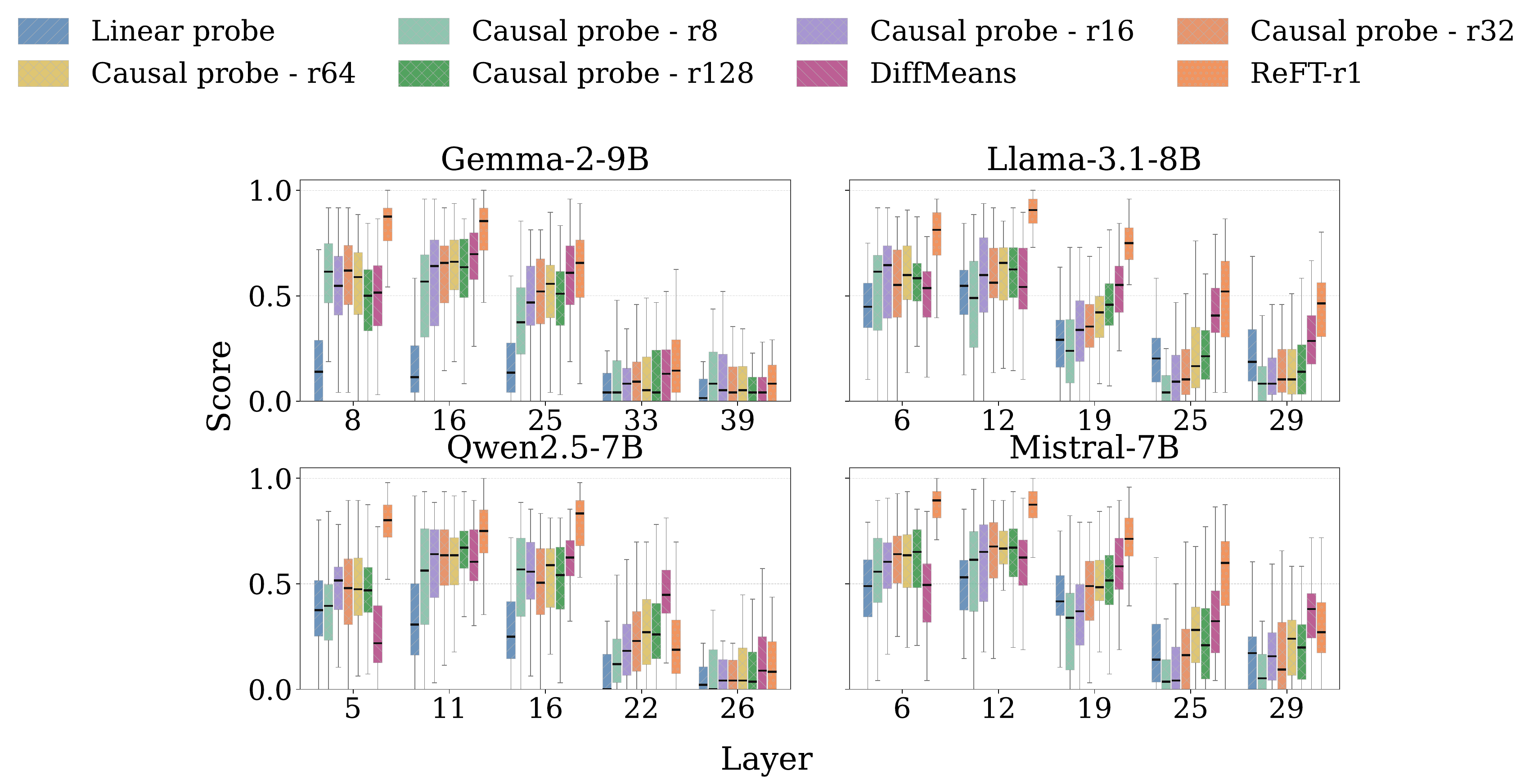}
        \caption{Overall steering score.}
        \label{fig:causal_probe_steering}
    \end{subfigure}
    
    \vspace{0.5em}
    
    \begin{subfigure}{0.49\linewidth}
        \centering
        \includegraphics[width=\linewidth]{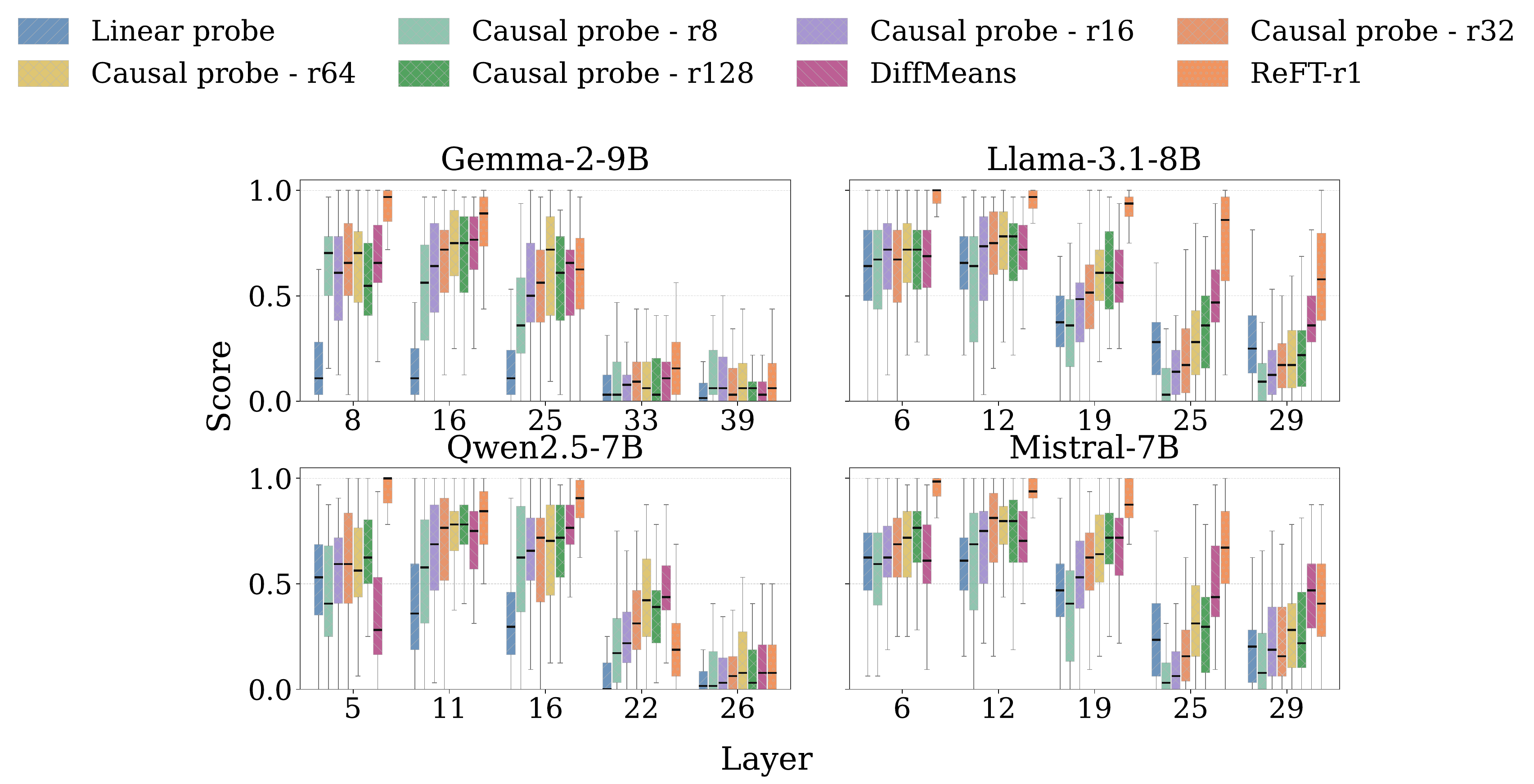}
        \caption{Concept score.}
        \label{fig:causal_probe_concept}
    \end{subfigure}
    \hfill
    \begin{subfigure}{0.49\linewidth}
        \centering
        \includegraphics[width=\linewidth]{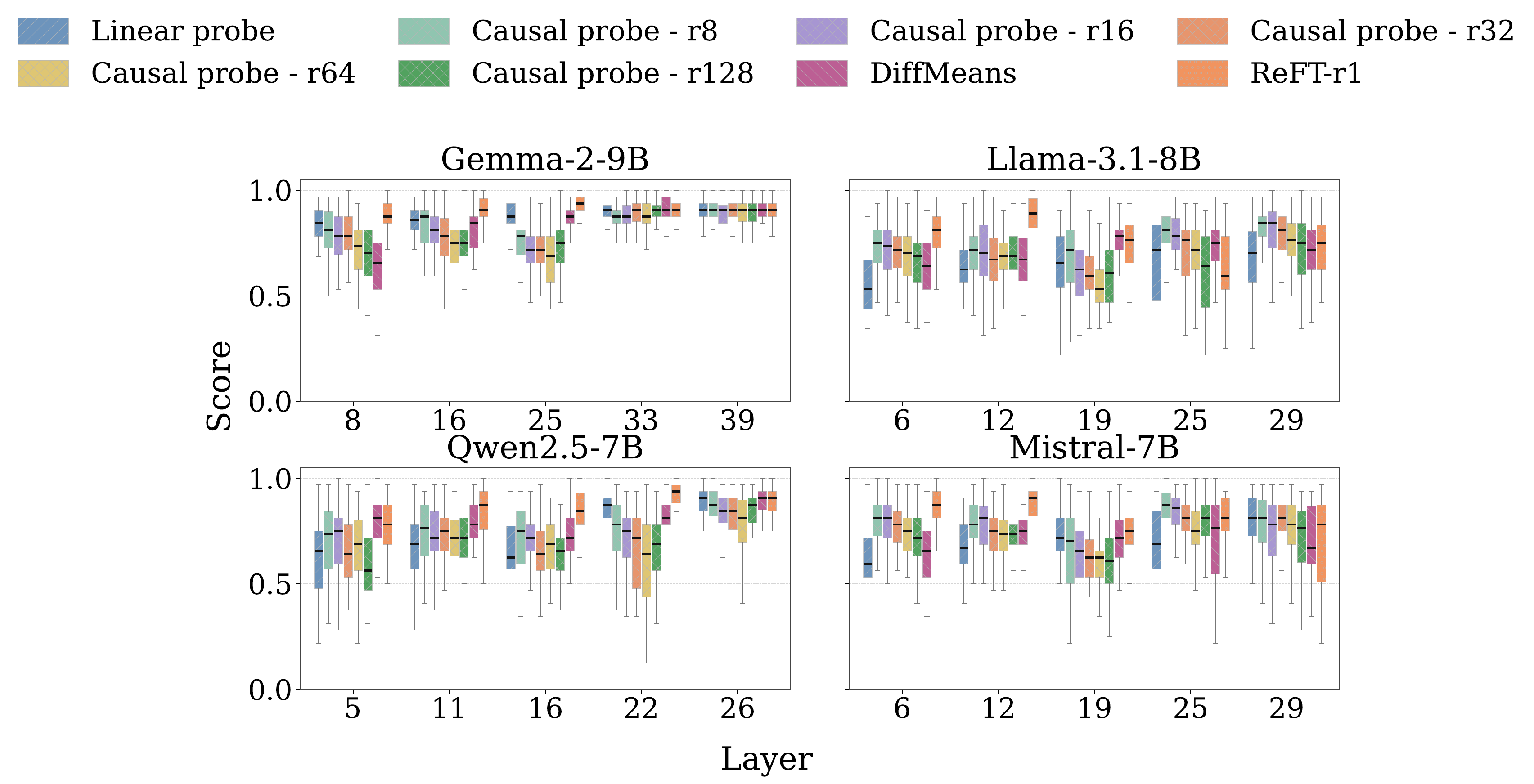}
        \caption{Fluency score.}
        \label{fig:causal_probe_fluency}
    \end{subfigure}
    
    \caption{
    \textbf{Causal probes improve steering primarily through improved concept expression.}
    We report the overall steering score and its concept and fluency components across models and layers.
    }
    \label{fig:full_causal_probe_scores}
\end{figure*}

Figure~\ref{fig:full_causal_probe_scores} reports the concept and fluency scores for all tested ranks of the causal probe. Causal probes substantially improve concept scores over unconstrained probes. Notably, models with higher-rank high-capacity subspaces continue to benefit as the causal-probe rank increases. This further supports our capacity findings: causal influence is concentrated in a low-dimensional subspace, and constraining the latent structure to this subspace improves its causal effectiveness. We observe no similarly strong trend in fluency. Causal probes are slightly more fluent on average, but most of their overall gain is explained by improved concept scores.

\end{document}